\pdfoutput=1
\documentclass[11pt]{article}

\usepackage[]{acl}

\usepackage{times}
\usepackage{latexsym}

\usepackage[T1]{fontenc}
\usepackage[utf8]{inputenc}

\usepackage{microtype}

\usepackage{booktabs} % \midrule \toprule
\usepackage{tabularx} % Table spanning entire page
\usepackage{hyperref}       % hyperlinks
\usepackage{url}            % simple URL typesetting
\usepackage{xurl} % Linebreak on - in URL
\usepackage{amsfonts}       % blackboard math symbols
\usepackage{nicefrac}       % compact symbols for 1/2, etc.
\usepackage{microtype}      % microtypography
\usepackage{xcolor}         % colors
\usepackage{graphicx}       % graphics
\usepackage{subfig} % Two figures side-by-side (Multilinguality)
\usepackage{pdflscape} % landscape page
\usepackage{changepage} % move table left via adjustwidth

\title{MTEB: Massive Text Embedding Benchmark}

\author{Niklas Muennighoff$^1$, Nouamane Tazi$^1$, Loïc Magne$^1$, Nils Reimers$^2$* \\
  $^1$Hugging Face \quad $^2$cohere.ai \\
  $^1${\tt firstname@hf.co} \quad $^2${\tt info@nils-reimers.de}\\
}

\begin{document}
\maketitle
\begingroup\def\thefootnote{*}\footnotetext{Most of the work done while at Hugging Face. Correspondence
to {\tt \href{mailto:n.muennighoff@gmail.com}{n.muennighoff@gmail.com}}.}\endgroup
\begin{abstract}

Text embeddings are commonly evaluated on a small set of datasets from a single task not covering their possible applications to other tasks. It is unclear whether state-of-the-art embeddings on semantic textual similarity (STS) can be equally well applied to other tasks like clustering or reranking. This makes progress in the field difficult to track, as various models are constantly being proposed without proper evaluation. To solve this problem, we introduce the Massive Text Embedding Benchmark (MTEB). MTEB spans 8 embedding tasks covering a total of 58 datasets and 112 languages. Through the benchmarking of 33 models on MTEB, we establish the most comprehensive benchmark of text embeddings to date. We find that no particular text embedding method dominates across all tasks. This suggests that the field has yet to converge on a universal text embedding method and scale it up sufficiently to provide state-of-the-art results on all embedding tasks. MTEB comes with open-source code and a public leaderboard at
\url{https://github.com/embeddings-benchmark/mteb}.

\end{abstract}

\section{Introduction}

Natural language embeddings power a variety of use cases from clustering and topic representation \cite{aggarwal2012survey, angelov2020top2vec} to search systems and text mining \cite{huang2020embedding, zhu2021bing, nayak2021google} to feature representations for downstream models \cite{saharia2022photorealistic, borgeaud2022improving}. Using generative language models or cross-encoders for these applications is often intractable, as they may require exponentially more computations \cite{reimers2019sentence}.

However, the evaluation regime of current text embedding models rarely covers the breadth of their possible use cases. For example, SimCSE \cite{gao2021simcse} or SBERT \cite{reimers2019sentence} solely evaluate on STS and classification tasks, leaving open questions about the transferability of the embedding models to search or clustering tasks. STS is known to poorly correlate with other real-world use cases \cite{neelakantan2022text, wang2021tsdae}. Further, evaluating embedding methods on many tasks requires implementing multiple evaluation pipelines. Implementation details like pre-processing or hyperparameters may influence the results making it unclear whether performance improvements simply come from a favorable evaluation pipeline. This leads to the ``blind'' application of these models to new use cases in industry or requires incremental work to reevaluate them on different tasks.

The Massive Text Embedding Benchmark (MTEB) aims to provide clarity on how models perform on a variety of embedding tasks and thus serves as the gateway to finding universal text embeddings applicable to a variety of tasks. MTEB consists of 58 datasets covering 112 languages from 8 embedding tasks: Bitext mining, classification, clustering, pair classification, reranking, retrieval, STS and summarization. MTEB software is available open-source\footnote{\url{https://github.com/embeddings-benchmark/mteb}} enabling evaluation of any embedding model by adding less than 10 lines of code. Datasets and the MTEB leaderboard are available on the Hugging Face Hub\footnote{\url{https://huggingface.co/spaces/mteb/leaderboard}}.

We evaluate over 30 models on MTEB with additional speed and memory benchmarking to provide a holistic view of the state of text embedding models. We cover both models available open-source as well as models accessible via APIs, such as the OpenAI Embeddings endpoint. We find there to be no single best solution, with different models dominating different tasks. Our benchmarking sheds light on the weaknesses and strengths of individual models, such as SimCSE's \cite{gao2021simcse} low performance on clustering and retrieval despite its strong performance on STS. We hope our work makes selecting the right embedding model easier and simplifies future embedding research.

\begin{figure*}[th]
    \centering
    \begin{center}
        \includegraphics[width=\textwidth]{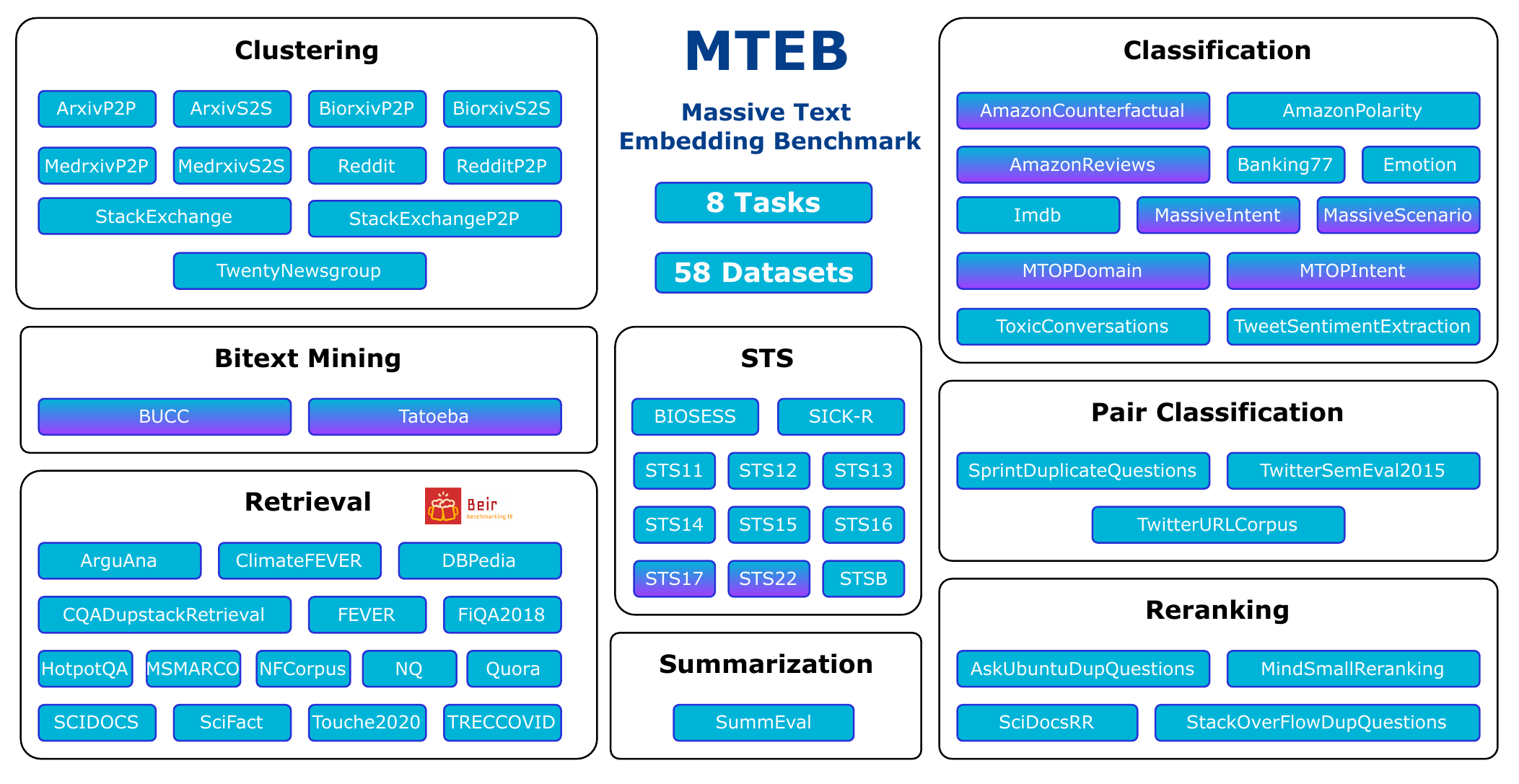}
        \caption{An overview of tasks and datasets in MTEB. Multilingual datasets are marked with a purple shade.}
        \label{fig:overview}
    \end{center}
\end{figure*}

\section{Related Work}

\begin{figure*}[t]
    \centering
    \begin{center}
        %\vspace{-7mm}
        \includegraphics[width=\textwidth]{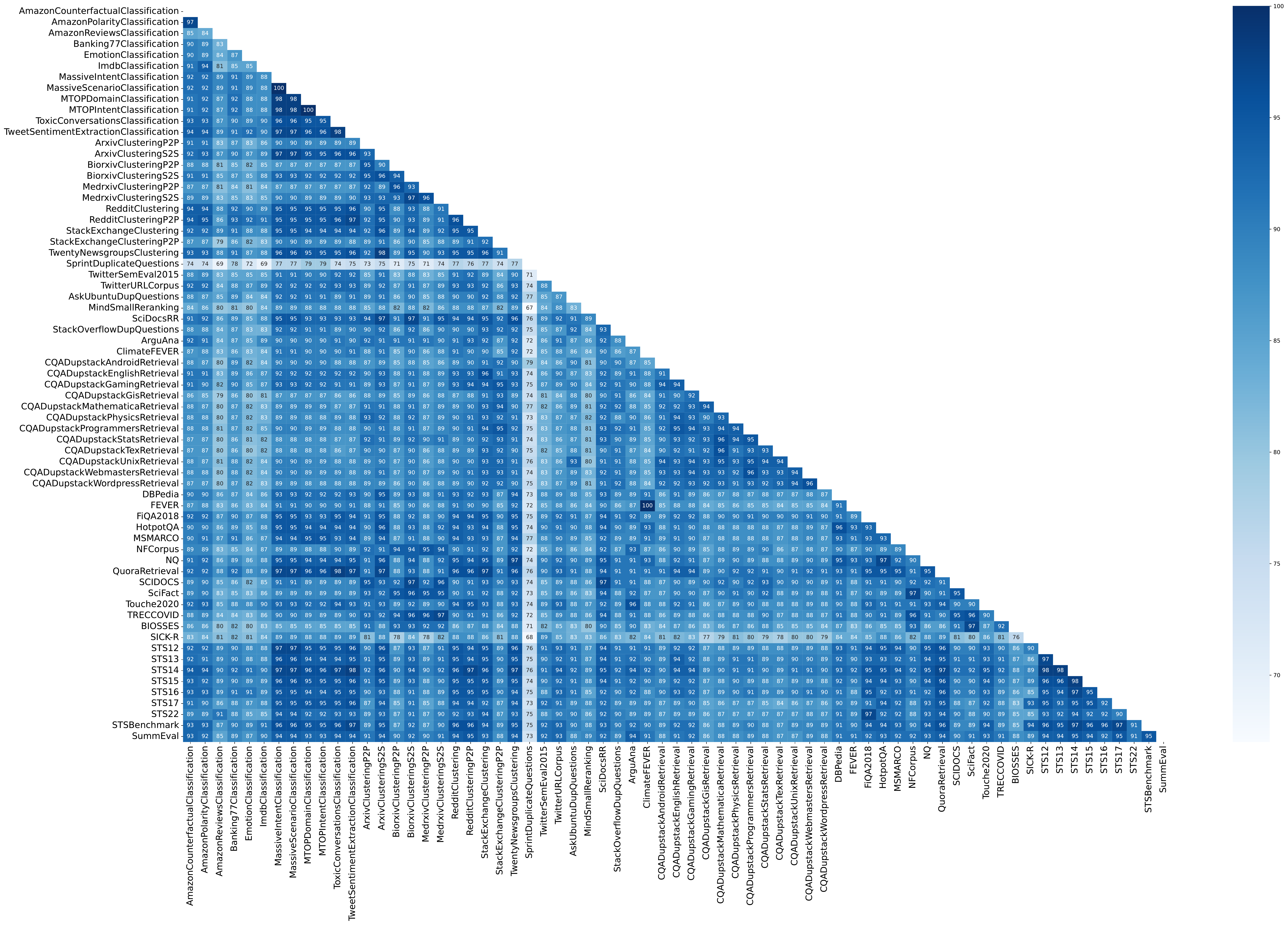}
        \caption{Similarity of MTEB datasets. We use the best model on MTEB STS (ST5-XXL, see Table~\ref{tab:results}) to embed 100 samples for each dataset. Cosine similarities between the averaged embeddings are computed and visualized.}
        \label{fig:similarity}
    \end{center}
\end{figure*}

\subsection{Benchmarks}

Benchmarks, such as (Super)GLUE~\cite{wang2018glue,wang2019superglue} or Big-BENCH~\cite{srivastava2022beyond}, and evaluation frameworks~\cite{gao2021framework} play a key role in driving NLP progress. Yearly released SemEval datasets \cite{agirre2012semeval, agirre2013sem, agirre2014semeval, agirre2015semeval, agirre2016semeval} are commonly used as the go-to benchmark for text embeddings. SemEval datasets correspond to the task of semantic textual similarity (STS) requiring models to embed similar sentences with geometrically close embeddings. Due to the limited expressivity of a single SemEval dataset, SentEval \cite{conneau2018senteval} aggregates multiple STS datasets. SentEval focuses on fine-tuning classifiers on top of embeddings. It lacks tasks like retrieval or clustering, where embeddings are directly compared without additional classifiers. Further, the toolkit was proposed in 2018 and thus does not provide easy support for recent trends like text embeddings from transformers \cite{reimers2019sentence}. Due to the insufficiency of STS benchmarking, USEB \cite{wang2021tsdae} was introduced consisting mostly of reranking tasks. Consequently, it does not cover tasks like retrieval or classification. Meanwhile, the recently released BEIR Benchmark \cite{beir} has become the standard for the evaluation of embeddings for zero-shot information retrieval.

MTEB unifies datasets from different embedding tasks into a common, accessible evaluation framework. MTEB incorporates SemEval datasets (STS11 - STS22) and BEIR alongside a variety of other datasets from various tasks to provide a holistic performance review of text embedding models.

\subsection{Embedding Models}

Text embedding models like Glove \cite{pennington2014glove} lack context awareness and are thus commonly labeled as Word Embedding Models. They consist of a layer mapping each input word to a vector often followed by an averaging layer to provide a final embedding invariant of input length. Transformers \cite{vaswani2017attention} inject context awareness into language models via self-attention and form the foundation of most recent embedding models. BERT \cite{devlin2018bert} uses the transformer architecture and performs large-scale self-supervised pre-training. The resulting model can directly be used to produce text embeddings via an averaging operation alike Glove. Building on InferSent \cite{conneau2017supervised}, SBERT \cite{reimers2019sentence} demonstrated it to be beneficial to perform additional fine-tuning of the transformer for competitive embedding performance. Most recent fine-tuned embedding models use a contrastive loss objective to perform supervised fine-tuning on positive and negative text pairs \cite{gao2021simcse, wang2021tsdae, ni2021large, muennighoff2022sgpt}. Due to the large variety of available pre-trained transformers \cite{wolf2020transformers}, there is an at least equally large variety of potential text embedding models to be explored. This leads to confusion about which model provides practitioners with the best performance for their embedding use case.

We benchmark both word embedding and transformer models on MTEB quantifying gains provided by often much slower context aware models.

\section{The MTEB Benchmark}

\subsection{Desiderata}
\label{sec:desiderata}

MTEB is built on a set of desiderata: \textbf{(a) Diversity:} MTEB aims to provide an understanding of the usability of embedding models in various use cases. The benchmark comprises 8 different tasks, with up to 15 datasets each. Of the 58 total datasets in MTEB, 10 are multilingual, covering 112 different languages. Sentence-level and paragraph-level datasets are included to contrast performance on short and long texts. \textbf{(b) Simplicity:} MTEB provides a simple API for plugging in any model that given a list of texts can produce a vector for each list item with a consistent shape. This makes it possible to benchmark a diverse set of models. \textbf{(c) Extensibility:} New datasets for existing tasks can be benchmarked in MTEB via a single file that specifies the task and a Hugging Face dataset name where the data has been uploaded \cite{lhoest2021datasets}. New tasks require implementing a task interface for loading the data and an evaluator for benchmarking. We welcome dataset, task or metric contributions from the community via pull requests to continue the development of MTEB. \textbf{(d) Reproducibility:} Through versioning at a dataset and software level, we aim to make it easy to reproduce results in MTEB. JSON files corresponding to all results available in this paper have been made available together with the MTEB benchmark\footnote{\url{https://huggingface.co/datasets/mteb/results}}.

\subsection{Tasks and Evaluation}
\label{sec:taskeval}

Figure \ref{fig:overview} provides an overview of tasks and datasets available in MTEB. Dataset statistics are available in Table \ref{tab:tasks}. The benchmark consists of the following 8 task types:

\paragraph{Bitext Mining} Inputs are two sets of sentences from two different languages. For each sentence in the first set, the best match in the second set needs to be found. The matches are commonly translations. The provided model is used to embed each sentence and the closest pairs are found via cosine similarity. F1 serves as the main metric for bitext mining. Accuracy, precision and recall are also computed.
    
\paragraph{Classification} A train and test set are embedded with the provided model. The train set embeddings are used to train a logistic regression classifier with 100 maximum iterations, which is scored on the test set. The main metric is accuracy with average precision and f1 additionally provided.

\paragraph{Clustering} Given a set of sentences or paragraphs, the goal is to group them into meaningful clusters. A mini-batch k-means model with batch size 32 and k equal to the number of different labels~\cite{scikit-learn} is trained on the embedded texts. The model is scored using v-measure \cite{vmeasure}. V-measure does not depend on the cluster label, thus the permutation of labels does not affect the score.

\paragraph{Pair Classification} A pair of text inputs is provided and a label needs to be assigned. Labels are typically binary variables denoting duplicate or paraphrase pairs. The two texts are embedded and their distance is computed with various metrics (cosine similarity, dot product, euclidean distance, manhattan distance). Using the best binary threshold accuracy, average precision, f1, precision and recall are computed. The average precision score based on cosine similarity is the main metric.

\paragraph{Reranking} Inputs are a query and a list of relevant and irrelevant reference texts. The aim is to rank the results according to their relevance to the query. The model is used to embed the references which are then compared to the query using cosine similarity. The resulting ranking is scored for each query and averaged across all queries. Metrics are mean MRR@k and MAP with the latter being the main metric.

\paragraph{Retrieval} Each dataset consists of a corpus, queries and a mapping for each query to relevant documents from the corpus. The aim is to find these relevant documents. The provided model is used to embed all queries and all corpus documents and similarity scores are computed using cosine similarity. After ranking the corpus documents for each query based on the scores, nDCG@k, MRR@k, MAP@k, precision@k and recall@k are computed for several values of $k$. nDCG@10 serves as the main metric. MTEB reuses datasets and evaluation from BEIR \cite{beir}.

\paragraph{Semantic Textual Similarity (STS)} Given a sentence pair the aim is to determine their similarity. Labels are continuous scores with higher numbers indicating more similar sentences. The provided model is used to embed the sentences and their similarity is computed using various distance metrics. Distances are benchmarked with ground truth similarities using Pearson and Spearman correlations. Spearman correlation based on cosine similarity serves as the main metric \cite{reimers2016task}.

\paragraph{Summarization} A set of human-written and machine-generated summaries are provided. The aim is to score the machine summaries. The provided model is first used to embed all summaries. For each machine summary embedding, distances to all human summary embeddings are computed. The closest score (e.g. highest cosine similarity) is kept and used as the model's score of a single machine-generated summary. Pearson and Spearman correlations with ground truth human assessments of the machine-generated summaries are computed. Like for STS, Spearman correlation based on cosine similarity serves as the main metric \cite{reimers2016task}.

\subsection{Datasets}
\label{sec:datasets}

\begin{table*}[t!]
    \centering
    \resizebox{0.9\textwidth}{!}{\begin{tabular}{l|ccccccccc}
    \toprule
 & Class. & Clust. & PairClass. & Rerank. & Retr. & STS & Summ. & Avg. \\
Num. Datasets ($\rightarrow$) & 12 & 11 & 3 & 4 & 15 & 10 & 1 & 56 \\
\midrule
\midrule
\multicolumn{9}{l}{\emph{Self-supervised methods}}
\\
\midrule
Glove & 57.29 & 27.73 & 70.92 & 43.29 & 21.62 & 61.85 & 28.87 & 41.97 \\
Komninos & 57.65 & 26.57 & 72.94 & 44.75 & 21.22 & 62.47 & 30.49 & 42.06 \\
BERT & 61.66 & 30.12 & 56.33 & 43.44 & 10.59 & 54.36 & 29.82 & 38.33 \\
SimCSE-BERT-unsup & 62.50 & 29.04 & 70.33 & 46.47 & 20.29 & 74.33 & 31.15 & 45.45 \\
\midrule
\multicolumn{9}{l}{\emph{Supervised methods}}
\\
\midrule
SimCSE-BERT-sup & 67.32 & 33.43 & 73.68 & 47.54 & 21.82 & 79.12 & 23.31 & 48.72 \\
coCondenser-msmarco & 64.71 & 37.64 & 81.74 & 51.84 & 32.96 & 76.47 & 29.50 & 52.35 \\
Contriever & 66.68 & 41.10 & 82.53 & 53.14 & 41.88 & 76.51 & 30.36 & 56.00 \\
SPECTER & 52.37 & 34.06 & 61.37 & 48.10 & 15.88 & 61.02 & 27.66 & 40.28 \\
LaBSE & 62.71 & 29.55 & 78.87 & 48.42 & 18.99 & 70.80 & 31.05 & 45.21 \\
LASER2 & 53.65 & 15.28 & 68.86 & 41.44 & 7.93 & 55.32 & 26.80 & 33.63 \\
MiniLM-L6 & 63.06 & 42.35 & 82.37 & 58.04 & 41.95 & 78.90 & 30.81 & 56.26 \\
MiniLM-L12 & 63.21 & 41.81 & 82.41 & \underline{58.44} & 42.69 & 79.80 & 27.90 & 56.53 \\
MiniLM-L12-multilingual & 64.30 & 37.14 & 78.45 & 53.62 & 32.45 & 78.92 & 30.67 & 52.44 \\
MPNet & 65.07 & \underline{43.69} & 83.04 & \textbf{59.36} & 43.81 & 80.28 & 27.49 & 57.78 \\
MPNet-multilingual & 67.91 & 38.40 & 80.81 & 53.80 & 35.34 & 80.73 & \textbf{31.57} & 54.71 \\
OpenAI Ada Similarity & 70.44 & 37.52 & 76.86 & 49.02 & 18.36 & 78.60 & 26.94 & 49.52 \\
SGPT-125M-nli & 61.46 & 30.95 & 71.78 & 47.56 & 20.90 & 74.71 & 30.26 & 45.97 \\
SGPT-5.8B-nli & 70.14 & 36.98 & 77.03 & 52.33 & 32.34 & 80.53 & 30.38 & 53.74 \\
SGPT-125M-msmarco & 60.72 & 35.79 & 75.23 & 50.58 & 37.04 & 73.41 & 28.90 & 51.23 \\
SGPT-1.3B-msmarco & 66.52 & 39.92 & 79.58 & 54.00 & 44.49 & 75.74 & 25.44 & 56.11 \\
SGPT-2.7B-msmarco & 67.13 & 39.83 & 80.65 & 54.67 & 46.54 & 76.83 & 27.87 & 57.12 \\
SGPT-5.8B-msmarco & 68.13 & 40.35 & 82.00 & 56.56 & \textbf{50.25} & 78.10 & 24.75 & 58.81 \\
SGPT-BLOOM-7.1B-msmarco & 66.19 & 38.93 & 81.90 & 55.65 & 48.21 & 77.74 & 24.99 & 57.44 \\
GTR-Base & 65.25 & 38.63 & 83.85 & 54.23 & 44.67 & 77.07 & 29.67 & 56.19 \\
GTR-Large & 67.14 & 41.60 & 85.33 & 55.36 & 47.42 & 78.19 & 29.50 & 58.28 \\
GTR-XL & 67.11 & 41.51 & \textbf{86.13} & 55.96 & 47.96 & 77.80 & 30.21 & 58.42 \\
GTR-XXL & 67.41 & 42.42 & \underline{86.12} & \underline{56.65} & \underline{48.48} & 78.38 & 30.64 & \underline{58.97} \\
ST5-Base & 69.81 & 40.21 & 85.17 & 53.09 & 33.63 & 81.14 & \underline{31.39} & 55.27 \\
ST5-Large & 72.31 & 41.65 & 84.97 & 54.00 & 36.71 & \underline{81.83} & 29.64 & 57.06 \\
ST5-XL & \underline{72.84} & 42.34 & 86.06 & 54.71 & 38.47 & 81.66 & 29.91 & 57.87 \\
ST5-XXL & \textbf{73.42} & \textbf{43.71} & 85.06 & 56.43 & 42.24 & \textbf{82.63} & 30.08 & \textbf{59.51} \\
    \bottomrule
    \end{tabular}}
    \caption{Average of the main metric (see Section \ref{sec:taskeval}) per task per model on MTEB English subsets.}
    \label{tab:results}
\end{table*}

To further the diversity of MTEB, datasets of varying text lengths are included. All datasets are grouped into three categories:

\paragraph{Sentence to sentence (S2S)} A sentence is compared with another sentence. An example of S2S are all current STS tasks in MTEB, where the similarity between two sentences is assessed. 

\paragraph{Paragraph to paragraph (P2P)} A paragraph is compared with another paragraph. MTEB imposes no limit on the input length, leaving it up to the models to truncate if necessary. Several clustering tasks are framed as both S2S and P2P tasks. The former only compare titles, while the latter include both title and content. For ArxivClustering, for example, abstracts are concatenated to the title in the P2P setting.

\paragraph{Sentence to paragraph (S2P)} A few retrieval datasets are mixed in a S2P setting. Here a query is a single sentence, while documents are long paragraphs consisting of multiple sentences.\\

Similarities across 56 MTEB datasets are visualized in Figure \ref{fig:similarity}. Several datasets rely on the same corpora, such as ClimateFEVER and FEVER, resulting in a score of 1. Clusters of similar datasets can be seen among CQADupstack variations and STS datasets. S2S and P2P variations of the same dataset tend to also be similar. Scientific datasets, such as SciDocsRR, SciFact, ArxivClustering, show high similarities among each other even when coming from different tasks (Reranking, Retrieval and Clustering in this case).

\section{Results}

\subsection{Models}

\begin{figure}[t]
    \centering
    \includegraphics[width=\linewidth]{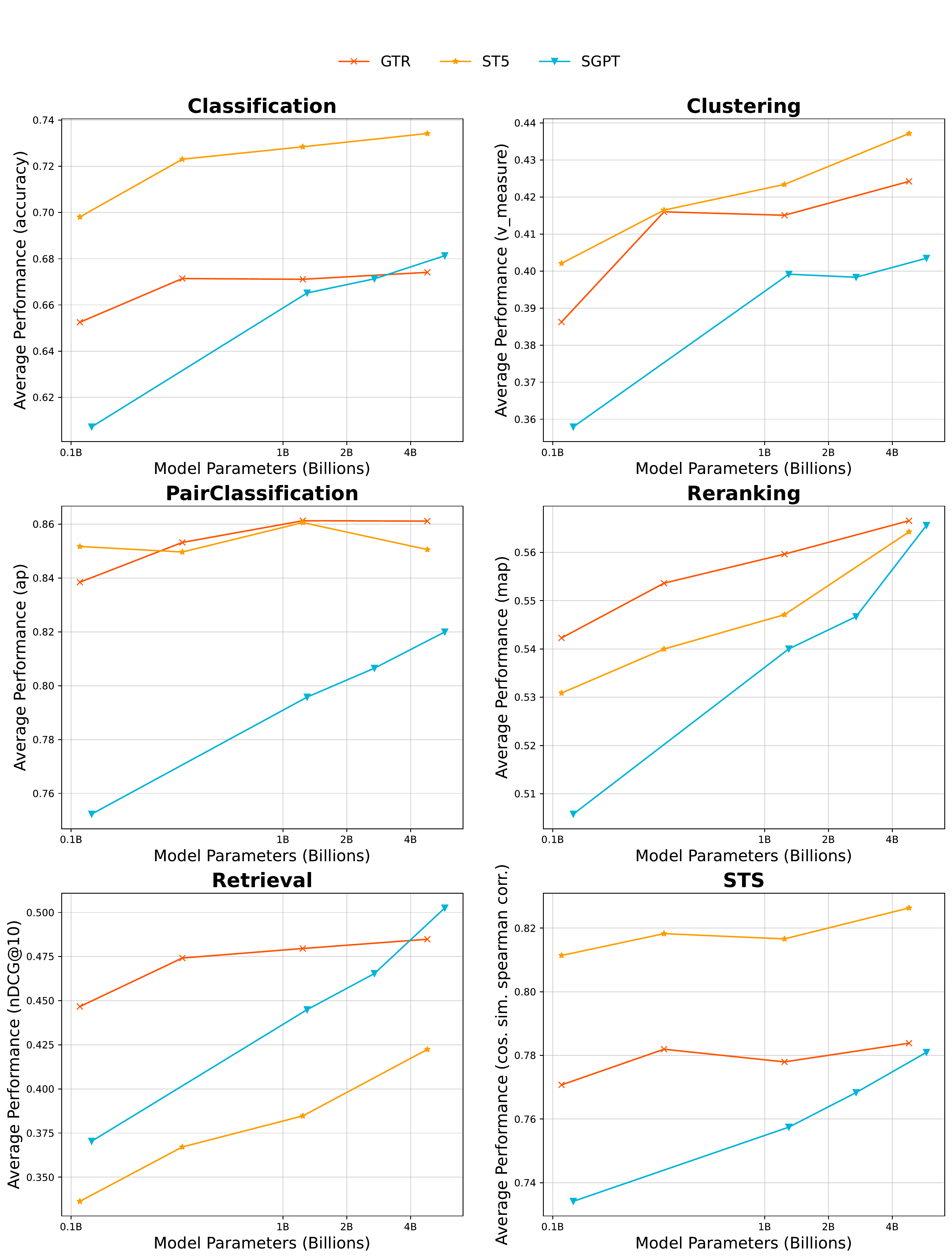}
    \caption{MTEB performance scales with model size. The smallest SGPT variant underperforms similar-sized GTR and ST5 variants. This may be due to the bias-only fine-tuning SGPT employs, which catches up with full fine-tuning only as model size and thus the number of bias parameters increases \cite{muennighoff2022sgpt}.}
    \label{fig:scale}
\end{figure}

We evaluate on the test splits of all datasets except for MSMARCO, where the dev split is used following~\citet{beir}. We benchmark models claiming state-of-the-art results on various embedding tasks leading to a high representation of transformers~\cite{vaswani2017attention}. We group models into self-supervised and supervised methods.

\paragraph{Self-supervised methods}

\begin{figure*}[t]
    \centering
    \begin{center}
        \includegraphics[width=\textwidth]{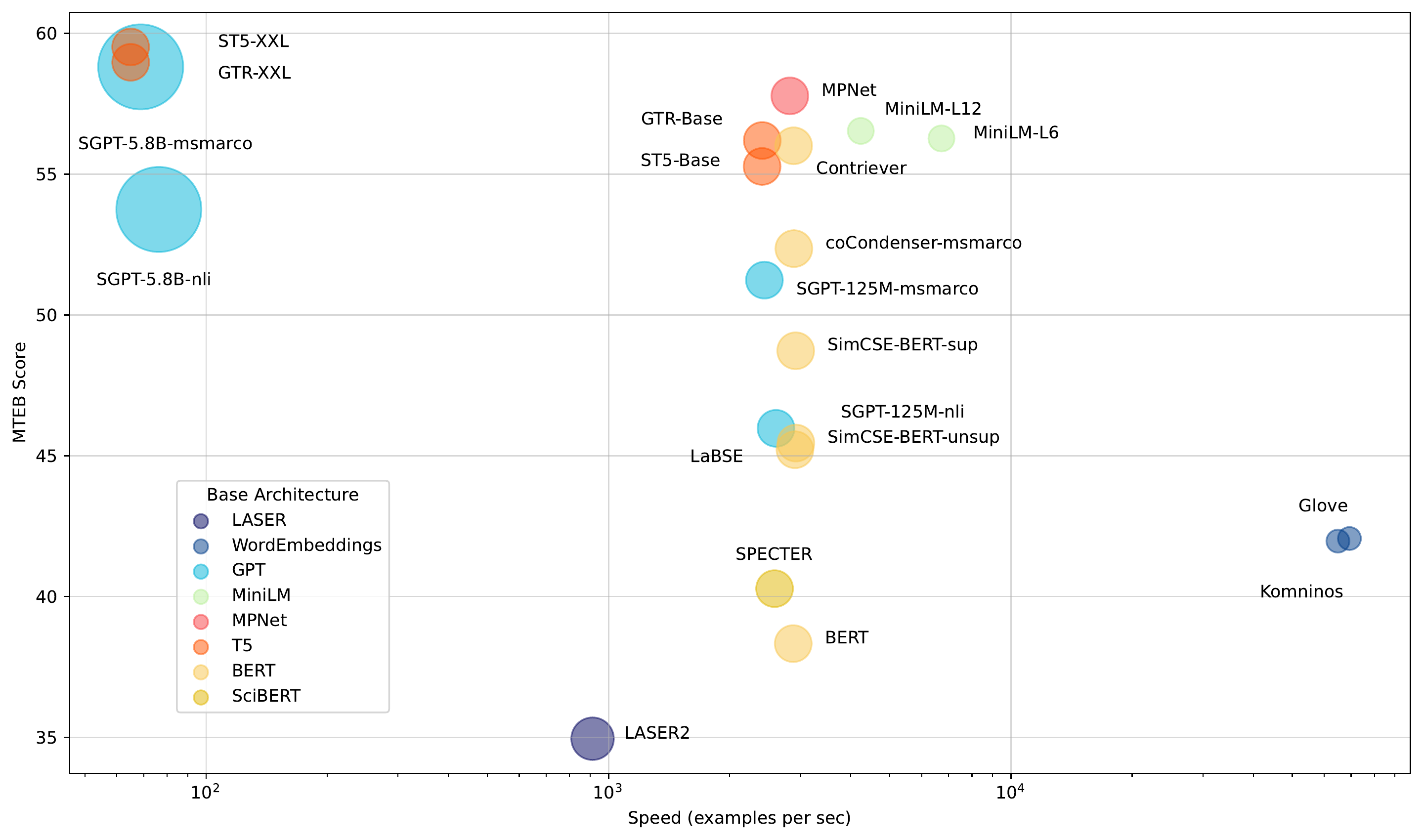}
        \caption{Performance, speed, and size of produced embeddings (size of the circles) of different embedding models. Embedding sizes range from 1.2 kB (Glove / Komninos) to 16.4 kB (SGPT-5.8B) per example. Speed was benchmarked on STS15 using 1x Nvidia A100 80GB with CUDA 11.6.}
        \label{fig:latency}
    \end{center}
\end{figure*}

\textbf{(a) Transformer-based} BERT \cite{devlin2018bert} is trained using self-supervised mask and sentence prediction tasks. By taking the mean across the sequence length (mean-pooling) the model can directly be used to produce text embeddings. SimCSE-Unsup \cite{gao2021simcse} uses BERT as a foundation and performs additional self-supervised training. \textbf{(b) Non-transformer}: Komninos \cite{komninos2016dependency} and Glove \cite{pennington2014glove} are two word embedding models that directly map words to vectors. Hence, their embeddings lack context awareness, but provide significant speed-ups.

\paragraph{Supervised methods} The original transformer model \cite{vaswani2017attention} consists of an encoder and decoder network. Subsequent transformers often train only encoders like BERT \cite{devlin2018bert} or decoders like GPT \cite{radford2019language}.

\textbf{(a) Transformer encoder methods} coCondenser \cite{gao2021unsupervised}, 
Contriever \cite{izacard2021towards}, LaBSE \cite{feng2020language} and SimCSE-BERT-sup \cite{gao2021simcse} are based on the pre-trained BERT model \cite{devlin2018bert}. coCondenser and Contriever add a self-supervised stage prior to supervised fine-tuning for a total of three training stages. LaBSE uses BERT to perform additional pre-training on parallel data to produce a competitive bitext mining model. SPECTER \cite{cohan2020specter} relies on the pre-trained SciBERT \cite{beltagy2019scibert} variant instead and fine-tunes on citation graphs. GTR \cite{ni2021large} and ST5 \cite{ni2021sentence} are based on the encoder part of the T5 model \cite{raffel2020exploring} and only differ in their fine-tuning datasets. After additional self-supervised training, ST5 does contrastive fine-tuning on NLI \cite{ni2021sentence, gao2021simcse} being geared towards STS tasks. Meanwhile, GTR fine-tunes on MSMARCO and focuses on retrieval tasks. MPNet and MiniLM correspond to fine-tuned embedding models \cite{reimers2019sentence} of the pre-trained MPNet \cite{song2020mpnet} and MiniLM \cite{wang2020minilm} models using diverse datasets to target any embedding use case. 

\textbf{(b) Transformer decoder methods} SGPT Bi-Encoders \cite{muennighoff2022sgpt} perform contrastive fine-tuning of <0.1\% of pre-trained parameters using weighted-mean pooling. Similar to ST5 and GTR, SGPT-nli models are geared towards STS, while SGPT-msmarco models towards retrieval. SGPT-msmarco models embed queries and documents for retrieval with different special tokens to help the model distinguish their role. For non-retrieval tasks, we use its query representations. We benchmark publicly available SGPT models based on GPT-NeoX \cite{gpt-neox}, GPT-J \cite{gpt-j} and BLOOM \cite{scao2022bloom}. Alternatively, cpt-text \cite{neelakantan2022text} passes pre-trained GPT decoders through a two-stage process using last token pooling to provide embeddings from decoders. We benchmark their models via the OpenAI Embeddings API\footnote{\url{https://beta.openai.com/docs/guides/embeddings}}.

\textbf{(c) Non-transformer} LASER \cite{heffernan2022bitext} is the only context aware non-transformer model we benchmark, relying on an LSTM \cite{hochreiter1997long} instead. Similar to LaBSE, the model trains on parallel data and focuses on bitext mining applications.

\subsection{Analysis}
\label{sec:analysis}

\begin{figure*}[t]
    \centering
    \subfloat[\centering Bitext Mining on Tatoeba]{{\includegraphics[width=0.95\textwidth]{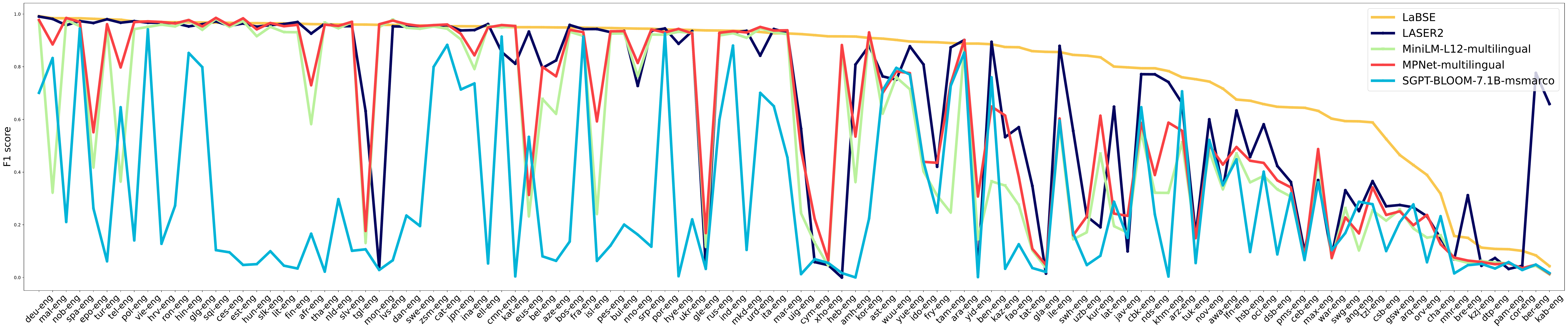}}}
    \qquad
    \subfloat[\centering Multilingual Classification]{{\includegraphics[width=0.45\textwidth]{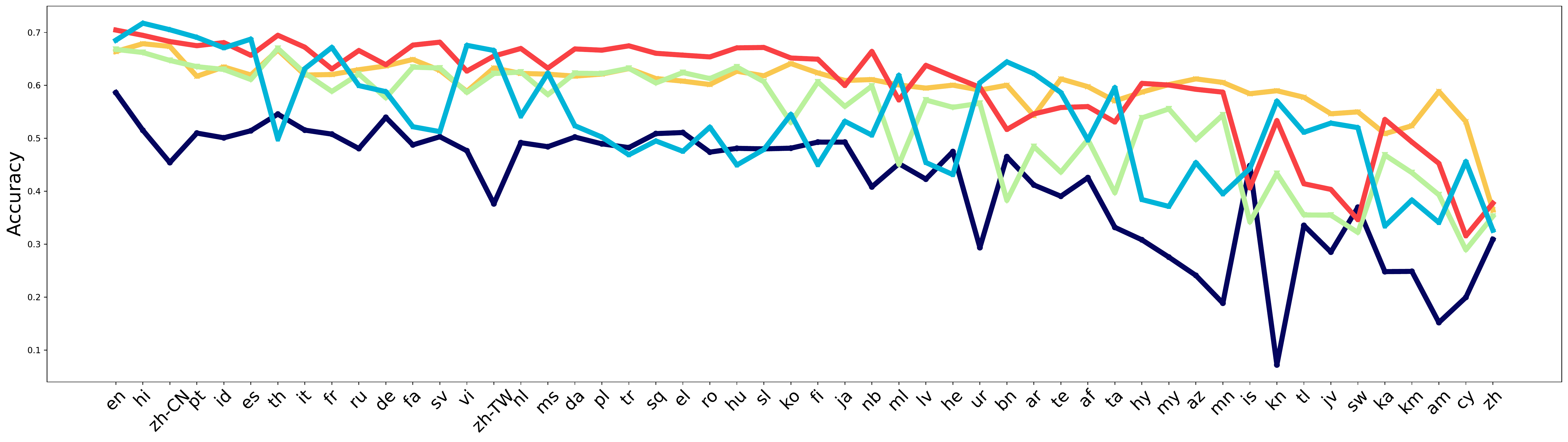}}}
    \qquad
    \subfloat[\centering Multi- and Crosslingual STS]{{\includegraphics[width=0.45\textwidth]{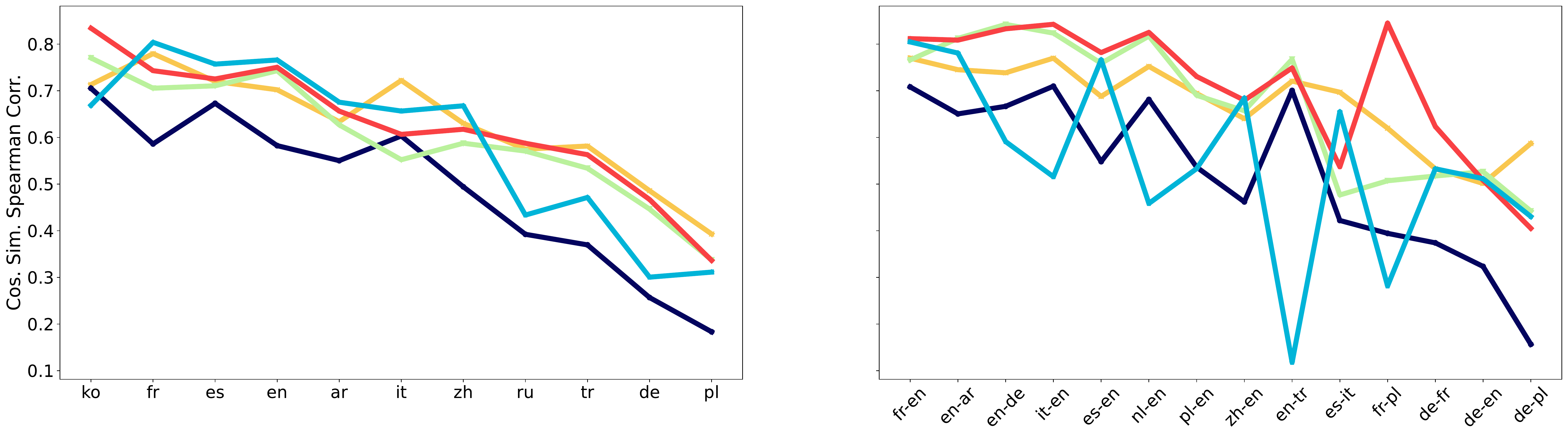}}}
    \caption{MTEB multilingual performance. Bitext mining is dominated by LaBSE, while classification and STS results are mixed. SGPT-BLOOM-7B1-msmarco tends to perform well on the languages BLOOM has been pre-trained on, such as Chinese, French and Portuguese.}
    \label{fig:multilingual}
\end{figure*}

Based on the results in Table \ref{tab:results}, we observe that there is considerable variability between tasks. No model claims the state-of-the-art in all seven English tasks. There is even more variability in the results per dataset present in the appendix. Further, there remains a large gap between self-supervised and supervised methods. Self-supervised large language models have been able to close this gap in many natural language generation tasks \cite{chowdhery2022palm}. However, they appear to still require supervised fine-tuning for competitive embedding performance.

We find that performance strongly correlates with model size, see Figure \ref{fig:scale}. A majority of MTEB tasks are dominated by multi-billion parameter models. However, these come at a significant cost as we investigate in Section \ref{sec:efficiency}.

\paragraph{Classification} ST5 models dominate the classification task across most datasets, as can be seen in detail in the full results in the appendix. ST5-XXL has the highest average performance, 3\% ahead of the best non-ST5 model, OpenAI Ada Similarity.

\paragraph{Clustering} Despite being almost 50x smaller, the MPNet embedding model is on par with the ST5-XXL state-of-the-art on Clustering. This may be due to the large variety of datasets MPNet (and MiniLM) has been fine-tuned on. Clustering requires coherent distances between a large number of embeddings. Models like SimCSE-sup or SGPT-nli, which are only fine-tuned on a single dataset, NLI, may produce incoherent embeddings when encountering topics unseen during fine-tuning. Relatedly, we find that the query embeddings of SGPT-msmarco and the Ada Search endpoint are competitive with SGPT-nli and the Ada Similarity endpoint, respectively. We refer to the public leaderboard\footnote{\url{https://huggingface.co/spaces/mteb/leaderboard}} for Ada Search results. This could be due to the MSMARCO dataset being significantly larger than NLI. Thus, while the OpenAI docs recommend using the similarity embeddings for clustering use cases\footnote{\url{https://beta.openai.com/docs/guides/embeddings/similarity-embeddings}}, the retrieval query embeddings may be the better choice in some cases.

\paragraph{Pair Classification} GTR-XL and GTR-XXL have the strongest performance. Pair classification is closest to STS in its framing, yet models rank significantly differently on the two tasks. This highlights the importance of benchmarking on a diverse set of tasks to avoid blindly reusing a model for a different task.

\paragraph{Reranking} MPNet and MiniLM models perform strongly on reranking tasks. On SciDocsRR \cite{cohan2020specter} they perform far better than bigger models, which is likely due to parts of SciDocsRR being included in their training data. Our scale of experiments and that of model pre-training make controlling for data contamination challenging. Thus, we ignore overlap of MTEB datasets with model training datasets in MTEB scores. As long as enough datasets are averaged, we believe these effects to be insignificant.

\paragraph{Retrieval} SGPT-5.8B-msmarco is the best embedding model on the BEIR subset in MTEB as well as on the full BEIR benchmark~\cite{beir, muennighoff2022sgpt}. The even larger 7.1B SGPT model making use of BLOOM~\cite{scao2022bloom} performs significantly weaker, which is likely due to the multilinguality of BLOOM. Models geared towards STS (SimCSE, ST5, SGPT-nli) perform badly on retrieval tasks. Retrieval tasks are unique in that there are two distinct types of texts: Queries and documents (``asymmetric''), while other tasks only have a single type of text (``symmetric''). On the QuoraRetrieval dataset, which has been shown to be largely symmetric \cite{muennighoff2022sgpt}, the playing field is more even with SGPT-5.8B-nli outperforming SGPT-5.8B-msmarco, see Table \ref{tab:addresults}.

\paragraph{STS \& Summarization} Retrieval models (GTR, SGPT-msmarco) perform badly on STS, while ST5-XXL has the highest performance. This highlights the bifurcation of the field into separate embedding models for retrieval (asymmetric) and similarity (symmetric) use cases \cite{muennighoff2022sgpt}.

\subsection{Efficiency}
\label{sec:efficiency}

We investigate the latency-performance trade-off of models in Figure \ref{fig:latency}. The graph allows for significant elimination of model candidates in the model selection process. It brings model selection down to three clusters: 

\paragraph{Maximum speed} Word Embedding models offer maximum speed with Glove taking the lead on both performance and speed, thus making the choice simple in this case.

\paragraph{Maximum performance} If latency is less important than performance, the left-hand side of the graph offers a cluster of highly performant, but slow models. Depending on the task at hand, GTR-XXL, ST5-XXL or SGPT-5.8B may be the right choice, see Section \ref{sec:analysis}. SGPT-5.8B comes with the additional caveat of its high-dimensional embeddings requiring more storage.

\paragraph{Speed and performance} The fine-tuned MPNet and MiniLM models lead the middle cluster making the choice easy.

\subsection{Multilinguality}

MTEB comes with 10 multilingual datasets across bitext mining, classification and STS tasks. We investigate performance on these in Figure \ref{fig:multilingual}. Tabular results can be found in Tables \ref{tab:addresultsmult}, \ref{tab:addresultsmultclf} and \ref{tab:addresultsmultsts}.

\paragraph{Bitext Mining} LaBSE \cite{feng2020language} performs strongly across a wide array of languages in bitext mining. Meanwhile, LASER2 shows high variance across different languages. While there are additional language-specific LASER2 models available for some of the languages we benchmark, we use the default multilingual LASER2 model for all languages. This is to provide a fair one-to-one comparison of models. In practice, however, the high variance of LASER2's performance may be resolved by mixing its model variants. MPNet, MiniLM and SGPT-BLOOM-7B1-msmarco perform poorly on languages they have not been pre-trained on, such as German for the latter.

\paragraph{Classification \& STS} On multilingual classification and STS, the multilingual MPNet provides the overall strongest performance. It outperforms the slightly faster multilingual MiniLM on almost all languages. Both models have been trained on the same languages, thus bringing decision-making down to performance vs speed. SGPT-BLOOM-7B1-msmarco provides state-of-the-art performance on languages like Hindi, Portuguese, Chinese or French, which the model has seen extensively during pre-training. It also performs competitively on languages like Russian or Japanese that unintentionally leaked into its pre-training data~\cite{muennighoff2022crosslingual}. However, it is not much ahead of the much cheaper MPNet. LASER2 performs consistently worse than other models.

\section{Conclusion}

In this work, we presented the Massive Text Embedding Benchmark (MTEB). Consisting of 8 text embedding tasks with up to 15 datasets each and covering 112 languages, MTEB aims to provide reliable embedding performance estimates. By open-sourcing MTEB alongside a leaderboard, we provide a foundation for further pushing the state-of-the-art of available text embeddings.

To introduce MTEB, we have conducted the most comprehensive benchmarking of text embeddings to date. Through the course of close to 5,000 experiments on over 30 different models, we have set up solid baselines for future research to build on. We found model performance on different tasks to vary strongly with no model claiming state-of-the-art on all tasks. Our studies on scaling behavior, model efficiency and multilinguality revealed various intricacies of models that should ease the decision-making process for future research or industry applications of text embeddings.

We welcome task, dataset or metric contributions to the MTEB codebase\footnote{\url{https://github.com/embeddings-benchmark/mteb}} as well as additions to the leaderboard via our automatic submission format\footnote{\url{https://huggingface.co/spaces/mteb/leaderboard}}.

\newpage

\section*{Acknowledgments}

This work was granted access to the HPC resources of Institut du d\'eveloppement et des ressources en informatique scientifique (IDRIS) du Centre national de la recherche scientifique (CNRS) under the allocation 2021-A0101012475 made by Grand \'equipement national de calcul intensif (GENCI). In particular, all the evaluations and data processing ran on the Jean Zay cluster of IDRIS, and we want to thank the IDRIS team for responsive support throughout the project, in particular R\'emi Lacroix.

We thank Douwe Kiela, Teven Le Scao and Nandan Thakur for feedback and suggestions.

\bibliography{custom}
\bibliographystyle{acl_natbib}

\newpage
\newpage

\appendix

\section{Datasets}
\label{sec:appdatasets}

Table \ref{tab:tasks} provides a summary along with statistics of all MTEB tasks. In the following, we give a brief description of each dataset included in MTEB.

\begin{table*}[!t]
    \centering
    \scriptsize
    \resizebox{\textwidth}{!}{\begin{tabular}{l|ccccccccc}
    \toprule
\bf Name & \bf Type & \bf Categ. & \bf \#Lang. & \bf Train & \bf Dev & \bf Test & \bf Train avg. & \bf Dev avg. & \bf Test avg. \\
& & & & \bf Samples & \bf Samples & \bf Samples & \bf chars & \bf chars & \bf chars \\
\midrule
\midrule
BUCC & BitextMining & s2s & 4 & 0 & 0 & 641684 & 0 & 0 & 101.3  \\
Tatoeba & BitextMining & s2s & 112 & 0 & 0 & 2000 & 0 & 0 & 39.4  \\
\midrule
AmazonCounterfactualClassification & Classification & s2s & 4 & 4018 & 335 & 670 & 107.3 & 109.2 & 106.1  \\
AmazonPolarityClassification & Classification & p2p & 1 & 3600000 & 0 & 400000 & 431.6 & 0 & 431.4  \\
AmazonReviewsClassification & Classification & s2s & 6 & 1200000 & 30000 & 30000 & 160.5 & 159.2 & 160.4  \\
Banking77Classification & Classification & s2s & 1 & 10003 & 0 & 3080 & 59.5 & 0 & 54.2  \\
EmotionClassification & Classification & s2s & 1 & 16000 & 2000 & 2000 & 96.8 & 95.3 & 96.6  \\
ImdbClassification & Classification & p2p & 1 & 25000 & 0 & 25000 & 1325.1 & 0 & 1293.8  \\
MassiveIntentClassification & Classification & s2s & 51 & 11514 & 2033 & 2974 & 35.0 & 34.8 & 34.6  \\
MassiveScenarioClassification & Classification & s2s & 51 & 11514 & 2033 & 2974 & 35.0 & 34.8 & 34.6  \\
MTOPDomainClassification & Classification & s2s & 6 & 15667 & 2235 & 4386 & 36.6 & 36.5 & 36.8  \\
MTOPIntentClassification & Classification & s2s & 6 & 15667 & 2235 & 4386 & 36.6 & 36.5 & 36.8  \\
ToxicConversationsClassification & Classification & s2s & 1 & 50000 & 0 & 50000 & 298.8 & 0 & 296.6  \\
TweetSentimentExtractionClassification & Classification & s2s & 1 & 27481 & 0 & 3534 & 68.3 & 0 & 67.8  \\
\midrule
ArxivClusteringP2P & Clustering & p2p & 1 & 0 & 0 & 732723 & 0 & 0 & 1009.9  \\
ArxivClusteringS2S & Clustering & s2s & 1 & 0 & 0 & 732723 & 0 & 0 & 74.0  \\
BiorxivClusteringP2P & Clustering & p2p & 1 & 0 & 0 & 75000 & 0 & 0 & 1666.2  \\
BiorxivClusteringS2S & Clustering & s2s & 1 & 0 & 0 & 75000 & 0 & 0 & 101.6  \\
MedrxivClusteringP2P & Clustering & p2p & 1 & 0 & 0 & 37500 & 0 & 0 & 1981.2  \\
MedrxivClusteringS2S & Clustering & s2s & 1 & 0 & 0 & 37500 & 0 & 0 & 114.7  \\
RedditClustering & Clustering & s2s & 1 & 0 & 420464 & 420464 & 0 & 64.7 & 64.7  \\
RedditClusteringP2P & Clustering & p2p & 1 & 0 & 0 & 459399 & 0 & 0 & 727.7  \\
StackExchangeClustering & Clustering & s2s & 1 & 0 & 417060 & 373850 & 0 & 56.8 & 57.0  \\
StackExchangeClusteringP2P & Clustering & p2p & 1 & 0 & 0 & 75000 & 0 & 0 & 1090.7  \\
TwentyNewsgroupsClustering & Clustering & s2s & 1 & 0 & 0 & 59545 & 0 & 0 & 32.0  \\
\midrule
SprintDuplicateQuestions & PairClassification & s2s & 1 & 0 & 101000 & 101000 & 0 & 65.2 & 67.9  \\
TwitterSemEval2015 & PairClassification & s2s & 1 & 0 & 0 & 16777 & 0 & 0 & 38.3  \\
TwitterURLCorpus & PairClassification & s2s & 1 & 0 & 0 & 51534 & 0 & 0 & 79.5  \\
\midrule
AskUbuntuDupQuestions & Reranking & s2s & 1 & 0 & 0 & 2255 & 0 & 0 & 52.5  \\
MindSmallReranking & Reranking & s2s & 1 & 231530 & 0 & 107968 & 69.0 & 0 & 70.9  \\
SciDocsRR & Reranking & s2s & 1 & 0 & 19594 & 19599 & 0 & 69.4 & 69.0  \\
StackOverflowDupQuestions & Reranking & s2s & 1 & 23018 & 3467 & 3467 & 49.6 & 49.8 & 49.8  \\
\midrule
ArguAna & Retrieval & p2p & 1 & 0 & 0 & 10080 & 0 & 0 & 1052.9  \\
ClimateFEVER & Retrieval & s2p & 1 & 0 & 0 & 5418128 & 0 & 0 & 539.1  \\
CQADupstackAndroidRetrieval & Retrieval & s2p & 1 & 0 & 0 & 23697 & 0 & 0 & 578.7  \\
CQADupstackEnglishRetrieval & Retrieval & s2p & 1 & 0 & 0 & 41791 & 0 & 0 & 467.1  \\
CQADupstackGamingRetrieval & Retrieval & s2p & 1 & 0 & 0 & 46896 & 0 & 0 & 474.7  \\
CQADupstackGisRetrieval & Retrieval & s2p & 1 & 0 & 0 & 38522 & 0 & 0 & 991.1  \\
CQADupstackMathematicaRetrieval & Retrieval & s2p & 1 & 0 & 0 & 17509 & 0 & 0 & 1103.7  \\
CQADupstackPhysicsRetrieval & Retrieval & s2p & 1 & 0 & 0 & 39355 & 0 & 0 & 799.4  \\
CQADupstackProgrammersRetrieval & Retrieval & s2p & 1 & 0 & 0 & 33052 & 0 & 0 & 1030.2  \\
CQADupstackStatsRetrieval & Retrieval & s2p & 1 & 0 & 0 & 42921 & 0 & 0 & 1041.0  \\
CQADupstackTexRetrieval & Retrieval & s2p & 1 & 0 & 0 & 71090 & 0 & 0 & 1246.9  \\
CQADupstackUnixRetrieval & Retrieval & s2p & 1 & 0 & 0 & 48454 & 0 & 0 & 984.7  \\
CQADupstackWebmastersRetrieval & Retrieval & s2p & 1 & 0 & 0 & 17911 & 0 & 0 & 689.8  \\
CQADupstackWordpressRetrieval & Retrieval & s2p & 1 & 0 & 0 & 49146 & 0 & 0 & 1111.9  \\
DBPedia & Retrieval & s2p & 1 & 0 & 4635989 & 4636322 & 0 & 310.2 & 310.1  \\
FEVER & Retrieval & s2p & 1 & 0 & 0 & 5423234 & 0 & 0 & 538.6  \\
FiQA2018 & Retrieval & s2p & 1 & 0 & 0 & 58286 & 0 & 0 & 760.4  \\
HotpotQA & Retrieval & s2p & 1 & 0 & 0 & 5240734 & 0 & 0 & 288.6  \\
MSMARCO & Retrieval & s2p & 1 & 0 & 8848803 & 8841866 & 0 & 336.6 & 336.8  \\
MSMARCOv2 & Retrieval & s2p & 1 & 138641342 & 138368101 & 0 & 341.4 & 342.0 & 0  \\
NFCorpus & Retrieval & s2p & 1 & 0 & 0 & 3956 & 0 & 0 & 1462.7  \\
NQ & Retrieval & s2p & 1 & 0 & 0 & 2684920 & 0 & 0 & 492.7  \\
QuoraRetrieval & Retrieval & s2s & 1 & 0 & 0 & 532931 & 0 & 0 & 62.9  \\
SCIDOCS & Retrieval & s2p & 1 & 0 & 0 & 26657 & 0 & 0 & 1161.9  \\
SciFact & Retrieval & s2p & 1 & 0 & 0 & 5483 & 0 & 0 & 1422.3  \\
Touche2020 & Retrieval & s2p & 1 & 0 & 0 & 382594 & 0 & 0 & 1720.1  \\
TRECCOVID & Retrieval & s2p & 1 & 0 & 0 & 171382 & 0 & 0 & 1117.4  \\
\midrule
BIOSSES & STS & s2s & 1 & 200 & 200 & 200 & 156.6 & 156.6 & 156.6  \\
SICK-R & STS & s2s & 1 & 19854 & 19854 & 19854 & 46.1 & 46.1 & 46.1  \\
STS12 & STS & s2s & 1 & 4468 & 0 & 6216 & 100.7 & 0 & 64.7  \\
STS13 & STS & s2s & 1 & 0 & 0 & 3000 & 0 & 0 & 54.0  \\
STS14 & STS & s2s & 1 & 0 & 0 & 7500 & 0 & 0 & 54.3  \\
STS15 & STS & s2s & 1 & 0 & 0 & 6000 & 0 & 0 & 57.7  \\
STS16 & STS & s2s & 1 & 0 & 0 & 2372 & 0 & 0 & 65.3  \\
STS17 & STS & s2s & 11 & 0 & 0 & 500 & 0 & 0 & 43.3  \\
STS22 & STS & p2p & 18 & 0 & 0 & 8060 & 0 & 0 & 1992.8  \\
STSBenchmark & STS & s2s & 1 & 11498 & 3000 & 2758 & 57.6 & 64.0 & 53.6  \\
\midrule
SummEval & Summarization & p2p & 1 & 0 & 0 & 2800 & 0 & 0 & 359.8 \\
    \bottomrule
    \end{tabular}}
    \caption{Tasks in MTEB}
    \label{tab:tasks}
\end{table*}

\subsection{Clustering}

\paragraph{ArxivClusteringS2S, ArxivClusteringP2P, BiorxivClusteringS2S, BiorxivClusteringP2P, MedrxivClusteringP2P, MedrxivClusteringS2S} These datasets are custom-made for MTEB using the public APIs from arXiv\footnote{\url{https://arxiv.org/help/api/}} and bioRxiv/medRxiv\footnote{\url{https://api.biorxiv.org/}}. For S2S datasets, the input text is simply the title of the paper, while for P2P the input text is the concatenation of the title and the abstract. The cluster labels are generated using categories given to the papers by humans. For bioRxiv and medRxiv this category is unique, but for arXiv multiple categories can be given to a single paper so we only use the first one. For bioRxiv and medRxiv there is only one level of category (e.g. biochemistry, genetics, microbiology, etc.) hence we only perform clustering based on that label. For arXiv there is a main category and secondary category: for example "cs.AI" means the main category is Computer Science and the sub-category is AI, math.AG means the main category is Mathematics and the sub-category is Algrebraic Geometry etc. Hence, we create three types of splits:

\paragraph{(a) Main category clustering} Articles are only clustered based on the main category (Math, Physics, Computer Science etc.). This split evaluates coarse clustering capacity of a model.

\paragraph{(b) Secondary category clustering within the same main category} Articles are clustered based on their secondary category, but within a given main category, for example only Math papers that need to be clustered into Algebraic Geometry, Functional Analysis, Numerical Analysis etc. This split evaluates fine-grained clustering capacity of a model, as differentiating some sub-categories can be very difficult.

\paragraph{(c) Secondary category clustering} Articles are clustered based on their secondary category for all main categories, so the labels can be Number Theory, Computational Complexity, Astrophysics of Galaxies etc. These splits evaluate fine-grained clustering capacity, as well as multi-scale capacities i.e. is a model able to both separate Maths from Physics as well as Probability from Algebraic Topology at the same time.

For every dataset, split and strategy, we select subsets of all labels and then sample articles from those labels. This yields splits with a varying amount and size of clusters.

\paragraph{RedditClustering} \cite{geigle2021clustering}: Clustering of titles from 199 subreddits. Clustering of 25 splits, each with 10-50 classes, and each class with 100 - 1000 sentences

\paragraph{RedditClusteringP2P} Dataset created for MTEB using available data from Reddit posts\footnote{\url{https://huggingface.co/datasets/sentence-transformers/reddit-title-body}}. The task consists of clustering the concatenation of title+post according to their subreddit. It contains 10 splits, with 10 and 100 clusters per split and 1,000 to 100,000 posts.

\paragraph{StackExchangeClustering} \cite{geigle2021clustering} Clustering of titles from 121 stackexchanges. Clustering of 25 splits, each with 10-50 classes, and each class with 100-1000 sentences.

\paragraph{StackExchangeClusteringP2P} Dataset created for MTEB using available data from StackExchange posts\footnote{\url{https://huggingface.co/datasets/flax-sentence-embeddings/stackexchange_title_body_jsonl}}. The task consists of clustering the concatenation of title and post according to their subreddit. It contains 10 splits, with 10 to 100 clusters and 5,000 to 10,000 posts per split.

\paragraph{TwentyNewsgroupsClustering\footnote{\url{https://scikit-learn.org/0.19/datasets/twenty_newsgroups.html}}} Clustering of the 20 Newsgroups dataset, given titles of article the goal is to find the newsgroup (20 in total). Contains 10 splits, each with 20 classes, with each split containing between 1,000 and 10,000 titles.

\subsection{Classification}

\paragraph{AmazonCounterfactual} \cite{oneill2021amazoncounterfactual} A collection of Amazon customer reviews annotated for counterfactual detection pair classification. For each review the label is either "counterfactual" or "not-counterfactual". This is a multilingual dataset with 4 available languages.

\paragraph{AmazonPolarity} \cite{mcauley2013amazon} A collection of Amazon customer reviews annotated for polarity classification. For each review the label is either "positive" or "negative".

\paragraph{AmazonReviews} \cite{mcauley2013amazon} A collection of Amazon reviews designed to aid research in multilingual text classification. For each review the label is the score given by the review between 0 and 4 (1-5 stars). This is a multilingual dataset with 6 available languages.

\paragraph{Banking77} \cite{casanueva2020banking77} Dataset composed of online banking queries annotated with their corresponding intents. For each user query the label is an intent among 77 intents like 'activate\_my\_card', 'apple\_pay', 'bank\_transfer', etc.

\paragraph{Emotion} \cite{saravia2018emotion} Dataset of English Twitter messages with six basic emotions: anger, fear, joy, love, sadness, and surprise.

\paragraph{Imdb} \cite{maas2011imdb} Large movie review dataset with labels being positive or negative.

\paragraph{MassiveIntent} \cite{fitzgerald2022massive} A collection of Amazon Alexa virtual assistant utterances annotated with the associated intent. For each user utterance the label is one of 60 intents like 'play\textunderscore music', 'alarm\textunderscore set', etc. This is a multilingual dataset with 51 available languages.

\paragraph{MassiveScenario} \cite{fitzgerald2022massive} A collection of Amazon Alexa virtual assistant utterances annotated with the associated intent. For each user utterance the label is a theme among 60 scenarios like 'music', 'weather', etc. This is a multilingual dataset with 51 available languages.

\paragraph{MTOPDomain / MTOPIntent} Multilingual sentence datasets from the MTOP \cite{li2020mtop} benchmark. We refer to their paper for details.

\paragraph{ToxicConversations} Dataset from Kaggle competition\footnote{\url{https://www.kaggle.com/competitions/jigsaw-unintended-bias-in-toxicity-classification}}. Collection of comments from the Civil Comments platform together with annotations if the comment is toxic or not.

\paragraph{TweetSentimentExtraction} Dataset from Kaggle competition\footnote{\url{https://www.kaggle.com/competitions/tweet-sentiment-extraction}}. Sentiment classification of tweets as neutral, positive or negative.

\subsection{Pair Classification}

\paragraph{SprintDuplicateQuestions} \cite{shah2018adversarial}: Collection of questions from the Sprint community. The goal is to classify a pair of sentences as duplicates or not.

\paragraph{TwitterSemEval2015} \cite{xu2015semeval} Paraphrase-Pairs of Tweets from the SemEval 2015 workshop. The goal is to classify a pair of tweets as paraphrases or not.

\paragraph{TwitterURLCorpus} \cite{lan2017sentential} Paraphrase-Pairs of Tweets. The goal is to classify a pair of tweets as paraphrases or not.

\subsection{Bitext Mining}

\paragraph{BUCC} \cite{zweigenbaum2016bucc1, zweigenbaum2017bucc2, zweigenbaum2018bucc3} BUCC provides big set of sentences ($\sim$ 10-70k each) for English, French, Russian, German and Chinese, along with associated pairs annotation. The annotated pairs here corresponds to a pairs of translated sentences, i.e. a sentence and its translation in the other language.

\paragraph{Tatoeba} \cite{tatoeba} Tatoeba provides sets of sentences (1000 sentences each) for 112 languages with annoated associated pairs. Each pair is one sentence and its translation in another language.

\subsection{Reranking}

\paragraph{AskUbuntuDupQuestions\footnote{\url{https://github.com/taolei87/askubuntu}}} Questions from AskUbuntu with manual annotations marking pairs of questions as similar or dissimilar.

\paragraph{MindSmall}\cite{wu2020mind} Large-scale English Dataset for News Recommendation Research. Ranking news article titles given the title of a news article. The idea is to recommend other news from the one you are reading.

\paragraph{SciDocsRR} \cite{cohan2020scidocs} Ranking of related scientific papers based on their title.

\paragraph{StackOverflowDupQuestions} \cite{liu2018linkso} Stack Overflow Duplicate Questions Task for questions with the tags Java, JavaScript and Python, ranking questions as duplicates or not.

\subsection{Semantic Textual Similarity (STS)}

\paragraph{STS12, STS13, STS14, STS15, STS16, STS17, STS22, STSBenchmark} \cite{agirre2012semeval, agirre2013sem}\footnote{\url{https://alt.qcri.org/semeval2014/task10/}}\footnote{\url{https://alt.qcri.org/semeval2015/task2/}}\footnote{\url{https://alt.qcri.org/semeval2016/task1/}}\footnote{\url{https://competitions.codalab.org/competitions/33835}} Original STS benchmark, with scores from 0 to 5. The selection of sentences includes text from image captions, news headlines and user forums. In total they contain between 1,000 and 20,000 sentences. STS12 - STS16 and STSBenchmark are monolingual english benchmarks. STS17 and STS22 contain crosslingual pairs of sentences, where the goal is to assess the similarity of two sentences in different languages. STS17 has 11 language pairs (among Korean, Arabic, English, French, German, Turkish, Spanish, Italian and Dutch) and STS22 has 18 language pairs (among Arabic, English, French, German, Turkish, Spanish, Polish, Italian, Russian and Chinese).

\paragraph{BIOSSES\footnote{\url{https://tabilab.cmpe.boun.edu.tr/BIOSSES/DataSet.html}}} Contains 100 sentence pairs from the biomedical field.

\paragraph{SICK-R} \cite{agirre2014semeval} Sentences Involving Compositional Knowledge (SICK) contains a large number of sentence pairs ($10\,0000$) that are lexically, syntactically and semantically rich.

\subsection{Summarization}

\paragraph{SummEval} \cite{fabbri2020summeval} Summaries generated by recent summarization models trained on CNN or DailyMail alongside human annotations.

\subsection{Retrieval}

We refer to the BEIR paper \cite{beir}, which contains description of each dataset. For MTEB, we include all publicly available datasets: \textbf{ArguAna, ClimateFEVER, CQADupstack, DBPedia, FEVER, FiQA2018, HotpotQA, MSMARCO, NFCorpus, NQ, Quora, SCIDOCS, SciFact, Touche2020, TRECCOVID}.

\section{Limitations of MTEB}

While MTEB aims to be a diverse benchmark to provide holistic performance reviews, the benchmark has its limitations. We list them here:

\paragraph{1. Long document datasets} MTEB covers multiple text lengths (S2S, P2P, S2P), but very long documents are still missing. The longest datasets in MTEB have a few hundred words, and longer text sizes could be relevant for use cases like retrieval.

\paragraph{2. Task imbalance} Tasks in MTEB have a different amount of datasets with summarization consisting of only a single dataset. This means MTEB average scores, which are computed over all datasets, are biased towards tasks with many datasets, notably retrieval, classification and clustering. As MTEB grows, we hope to add more datasets to currently underrepresented tasks like summarization or pair classification.

\paragraph{3. Multinguality} MTEB contains multilingual classification, STS and bitext mining datasets. However, retrieval and clustering are English-only. SGPT-BLOOM-7B1-msmarco is geared towards multilingual retrieval datasets and due to the lack thereof cannot be comprehensively benchmarked in MTEB. Further, MTEB does not contain any code datasets that could be used to benchmark code models~\cite{neelakantan2022text,allal2023santacoder}. It should be easy to extend MTEB with datasets, such as CodeSearchNet~\cite{husain2019codesearchnet}, TyDI QA~\cite{clark2020tydi}, XOR QA~\cite{asai2020xor} or MIRACL~\cite{zhang2022making}. 

\paragraph{4. Additional modalities} Text embeddings are commonly used as input features for downstream models, such as in our classification task. This can involve other modalities, notably image content~\cite{carvalho2018cross,tan2019lxmert,muennighoff2020vilio,nichol2021glide,saharia2022photorealistic,weinbach2022m}. We have focused solely on natural language applications and leave extensive benchmarking of text embeddings as inputs for other modalities to future work.

\section{Examples}
\label{sec:examples}

Tables \ref{tab:classification_examples}-\ref{tab:summarization_examples} provide examples for each dataset for each task. For retrieval datasets, we refer to the BEIR paper \cite{beir}.

\begin{table*}[!t]
    \centering
    \tiny
    \resizebox{\textwidth}{!}{\begin{tabular}{l|l|l}
    \toprule
\multicolumn{1}{l|}{\textbf{Dataset}} & \multicolumn{1}{c}{\textbf{Text}} & \multicolumn{1}{|l}{\textbf{Label}} \\
\midrule
\midrule
AmazonCounterfactualClassification & \multicolumn{1}{p{10cm}|}{In person it looks as though it would have cost a lot more.} & counterfactual \\
\midrule
AmazonPolarityClassification & \multicolumn{1}{p{10cm}|}{an absolute masterpiece I am quite sure any of you actually taking the time to read this have played the game at least once, and heard at least a few of the tracks here. And whether you were aware of it or not, Mitsuda's music contributed greatly to the...} & positive \\
\midrule
AmazonReviewsClassification & \multicolumn{1}{p{10cm}|}{solo llega una unidad cuando te obligan a comprar dos Te obligan a comprar dos unidades y te llega solo una y no hay forma de reclamar, una autentica estafa, no compreis!!} & 0 \\
\midrule
Banking77Classification & \multicolumn{1}{p{10cm}|}{What currencies is an exchange rate calculated in?} & exchange\textunderscore rate \\
\midrule
EmotionClassification & \multicolumn{1}{p{10cm}|}{i feel so inhibited in someone elses kitchen like im painting on someone elses picture} & sadness \\
\midrule
ImdbClassification & \multicolumn{1}{p{10cm}|}{When I first saw a glimpse of this movie, I quickly noticed the actress who was playing the role of Lucille Ball. Rachel York's portrayal of Lucy is absolutely awful. Lucille Ball was an astounding comedian with incredible talent. To think about a legend like Lucille Ball being portrayed the way she was in the movie is horrendous. I cannot believe...} & negative \\
\midrule
MassiveIntentClassification & \multicolumn{1}{p{10cm}|}{réveille-moi à neuf heures du matin le vendredi} & alarm\textunderscore set \\
\midrule
MassiveScenarioClassification & \multicolumn{1}{p{10cm}|}{tell me the artist of this song} & music \\
\midrule
MTOPDomainClassification & \multicolumn{1}{p{10cm}|}{Maricopa County weather forecast for this week} & weather \\
\midrule
MTOPIntentClassification & \multicolumn{1}{p{10cm}|}{what ingredients do is have left} & GET\textunderscore INFO\textunderscore RECIPES\\
\midrule
ToxicConversationsClassification & \multicolumn{1}{p{10cm}|}{The guy's a damn cop, so what do you expect?} & toxic \\
\midrule
TweetSentimentExtractionClassification & \multicolumn{1}{p{10cm}|}{I really really like the song Love Story by Taylor Swift} & positive \\
    \bottomrule
    \end{tabular}}
    \caption{Classification examples}
    \label{tab:classification_examples}
\end{table*}

\begin{table*}[!t]
    \centering
    \tiny
    \resizebox{\textwidth}{!}{\begin{tabular}{l|l|l}
    \toprule
\multicolumn{1}{l|}{\textbf{Dataset}} & \multicolumn{1}{c}{\textbf{Text}} & \multicolumn{1}{|l}{\textbf{Cluster}} \\
\midrule
\midrule
ArxivClusteringP2P & \multicolumn{1}{p{10cm}|}{Finite groups of rank two which do not involve $Qd(p)$.
Let $p>3$ be a prime. We show that if $G$ is a finite group with $p$-rank equal to 2, then $G$ involves $Qd(p)$ if and only if $G$ $p'$-involves $Qd(p)$. This allows us to use a version of Glauberman's ZJ-theorem to give a more direct construction of finite group actions on mod-$p$ homotopy spheres. We give an example to illustrate that the above conclusion does not hold for $p \leq 3$.} & math \\
\midrule
ArxivClusteringS2S & \multicolumn{1}{p{10cm}|}{Vertical shift and simultaneous Diophantine approximation on polynomial curves} &  math\\
\midrule
BiorxivClusteringP2P & \multicolumn{1}{p{10cm}|}{Innate Immune sensing of Influenza A viral RNA through IFI16 promotes pyroptotic cell death Programmed cell death pathways are triggered by various stresses or stimuli, including viral infections. The mechanism underlying the regulation of these pathways upon Influenza A virus IAV infection is not well characterized. We report that a cytosolic DNA sensor IFI16 is...} & immunology \\
\midrule
BiorxivClusteringS2S & \multicolumn{1}{p{10cm}|}{Association of CDH11 with ASD revealed by matched-gene co-expression analysis and mouse behavioral} & neuroscience \\
\midrule
MedrxivClusteringP2P & \multicolumn{1}{p{10cm}|}{Temporal trends in the incidence of haemophagocytic lymphohistiocytosis: a nationwide cohort study from England 2003-2018. Haemophagocytic lymphohistiocytosis (HLH) is rare, results in high mortality and is increasingly being diagnosed. Little is known about what is driving the apparent rise in the incidence of this disease. Using national linked electronic health data from hospital admissions and death certification cases of HLH that were diagnosed in England between 1/1/2003 and 31/12/2018 were identified using a previously validated approach. We calculated incidence...} & infectious diseases \\
\midrule
MedrxivClusteringS2S & \multicolumn{1}{p{10cm}|}{Current and Lifetime Somatic Symptom Burden Among Transition-aged Young Adults on the Autism Spectrum} & psychiatry and clinical psychology \\
\midrule
RedditClustering & \multicolumn{1}{p{10cm}|}{Could anyone tell me what breed my bicolor kitten is?} & r/cats \\
\midrule
RedditClusteringP2P & \multicolumn{1}{p{10cm}|}{Headaches after working out? Hey guys! I’ve been diagnosed with adhd since I was seven. I just recently got rediagnosed (22f) and I’ve been out on a different medication, adderall I was normally taking vyvanse but because of cost and no insurance adderall was more affordable. I’ve noticed that if I take adderall and workout... } & r/ADHD \\
\midrule
StackExchangeClustering & \multicolumn{1}{p{10cm}|}{Does this property characterize a space as Hausdorff?} & math.stackexchange.com \\
\midrule
StackExchangeClusteringP2P & \multicolumn{1}{p{10cm}|}{Google play services error DEBUG: Application is pausing, which disconnects the RTMP client. I am having this issue from past day with Google Play Services Unity. What happens is, when I install app directly ot device via Unity, the Google Play Services work fine but when I upload it as beta to play store console and install it via that then it starts to give " DEBUG: Application is pausing, which disconnects the RTMP client" error. I have a proper SHA1 key.} & unity \\
\midrule
TwentyNewsgroupsClustering & \multicolumn{1}{p{10cm}|}{Commercial mining activities on the moon} & 14 \\ \bottomrule
    \end{tabular}}
    \caption{Clustering examples}
    \label{tab:clustering_examples}
\end{table*}

\begin{table*}[!t]
    \centering
    \tiny
    \resizebox{\textwidth}{!}{\begin{tabular}{l|l|l|l}
    \toprule
\multicolumn{1}{l|}{\textbf{Dataset}} & \multicolumn{1}{c}{\textbf{Sentence 1}} & \multicolumn{1}{|c}{\textbf{Sentence 2}} & \multicolumn{1}{l}{\textbf{Label}}\\
\midrule
\midrule
SprintDuplicateQuestions & \multicolumn{1}{p{5cm}|}{Franklin U722 USB modem signal strength} & \multicolumn{1}{p{5cm}|}{How do I know if my Franklin U772 USB Modem has a weak signal ?} & 1 \\
\midrule
TwitterSemEval2015 & \multicolumn{1}{p{5cm}|}{All the home alones watching 8 mile","All the home alones watching 8 mile} & \multicolumn{1}{p{5cm}|}{The last rap battle in 8 Mile nevr gets old ahah} & 0 \\
\midrule
TwitterURLCorpus & \multicolumn{1}{p{5cm}|}{How the metaphors we use to describe discovery affect men and women in the sciences} & \multicolumn{1}{p{5cm}|}{Light Bulbs or Seeds ? How Metaphors for Ideas Influence Judgments About Genius} & 0 \\
    \bottomrule
    \end{tabular}}
    \caption{Pair classification examples. Labels are binary.}
    \label{tab:pair_classification_examples}
\end{table*}

\begin{table*}[!t]
    \centering
    \tiny
    \resizebox{\textwidth}{!}{\begin{tabular}{l|l|l|l}
    \toprule
\multicolumn{1}{l|}{\textbf{Dataset}} & \multicolumn{1}{c}{\textbf{Query}} & \multicolumn{1}{|c}{\textbf{Positive}} & \multicolumn{1}{|c}{\textbf{Negative}} \\
\midrule
\midrule
AskUbuntuDupQuestions & \multicolumn{1}{p{5cm}|}{change the application icon theme but not changing the panel icons} & \multicolumn{1}{p{5cm}|}{change folder icons in ubuntu-mono-dark theme} & \multicolumn{1}{p{5cm}}{change steam tray icon back to default} \\
\midrule
% I found examples on mindsmall to not make much sense
MindSmallReranking & \multicolumn{1}{p{5cm}|}{Man accused in probe of Giuliani associates is freed on bail} & \multicolumn{1}{p{5cm}|}{Studies show these are the best and worst states for your retirement} & \multicolumn{1}{p{5cm}}{There are 14 cheap days to fly left in 2019: When are they and what deals can you score?} \\
\midrule
SciDocsRR & \multicolumn{1}{p{5cm}|}{Discovering social circles in ego networks} & \multicolumn{1}{p{5cm}|}{Benchmarks for testing community detection algorithms on directed and weighted graphs with overlapping communities.} & \multicolumn{1}{p{5cm}}{Improving www proxies performance with greedy-dual-size-frequency caching policy} \\
\midrule
StackOverflowDupQuestions & \multicolumn{1}{p{5cm}|}{Java launch error selection does not contain a main type} & \multicolumn{1}{p{5cm}|}{Error: Selection does not contain a main type} & \multicolumn{1}{p{5cm}}{Selection Sort in Java} \\
    \bottomrule
    \end{tabular}}
    \caption{Reranking examples}
    \label{tab:reranking_examples}
\end{table*}

\begin{table*}[!t]
    \centering
    \tiny
    \resizebox{\textwidth}{!}{\begin{tabular}{l|l|l|l}
    \toprule
\multicolumn{1}{l|}{\textbf{Dataset}} & \multicolumn{1}{c}{\textbf{Sentence 1}} & \multicolumn{1}{|c}{\textbf{Sentence 2}} & \multicolumn{1}{|c}{\textbf{Score}} \\
\midrule
\midrule
BIOSSES & \multicolumn{1}{p{5cm}|}{It has recently been shown that Craf is essential for Kras G12D-induced NSCLC.} & \multicolumn{1}{p{5cm}|}{It has recently become evident that Craf is essential for the onset of Kras-driven non-small cell lung cancer.} & 4.0 \\
\midrule
SICK-R & \multicolumn{1}{p{5cm}|}{A group of children is playing in the house and there is no man standing in the background} & \multicolumn{1}{p{5cm}|}{A group of kids is playing in a yard and an old man is standing in the background} & 3.2 \\
\midrule
STS12 & \multicolumn{1}{p{5cm}|}{Nationally, the federal Centers for Disease Control and Prevention recorded 4,156 cases of West Nile, including 284 deaths.} & \multicolumn{1}{p{5cm}|}{There were 293 human cases of West Nile in Indiana in 2002, including 11 deaths statewide.} & 1.7 \\
\midrule
STS13 & \multicolumn{1}{p{5cm}|}{this frame has to do with people ( the residents ) residing in locations , sometimes with a co-resident .} & \multicolumn{1}{p{5cm}|}{inhabit or live in ; be an inhabitant of ;} & 2.8 \\
\midrule
STS14 & \multicolumn{1}{p{5cm}|}{then the captain was gone.} & \multicolumn{1}{p{5cm}|}{then the captain came back.} & 0.8  \\
\midrule
STS15 & \multicolumn{1}{p{5cm}|}{you 'll need to check the particular policies of each publisher to see what is allowed and what is not allowed.} & \multicolumn{1}{p{5cm}|}{if you need to publish the book and you have found one publisher that allows it.} & 3.0 \\
\midrule
STS16 & \multicolumn{1}{p{5cm}|}{you do not need to worry.} & \multicolumn{1}{p{5cm}|}{you don 't have to worry.} & 5.0 \\
\midrule
STS17 & \multicolumn{1}{p{5cm}|}{La gente muestra su afecto el uno por el otro.} & \multicolumn{1}{p{5cm}|}{A women giving something to other lady.} & 1.4 \\
\midrule
STS22 & \multicolumn{1}{p{5cm}|}{El secretario general de la Asociación Gremial de los Trabajadores del Subte y Premetro de Metrodelegados, Beto Pianelli, dijo que el Gobierno porteño debe convocar “inmediatamente” a licitación para la compra de nuevos trenes y retirar los que quedan en circulación...} & \multicolumn{1}{p{5cm}|}{En diálogo con el servicio informativo de la Radio Pública, el ministro de Salud de la Nación, Ginés González García, habló sobre el avance del coronavirus en la Argentina y se manifestó a favor de prorrogar la cuarentena obligatoria dispuesta por...} & 1 \\
\midrule
STSBenchmark & \multicolumn{1}{p{5cm}|}{A man is playing the cello.} & \multicolumn{1}{p{5cm}|}{A man seated is playing the cello.} & 4.25 \\
\midrule
    \bottomrule
    \end{tabular}}
    \caption{STS examples. Scores are continuous between 0 and 5 (included).}
    \label{tab:sts_examples}
\end{table*}

\begin{table*}[!t]
    \centering
    \tiny
    \resizebox{\textwidth}{!}{\begin{tabular}{l|l|l}
    \toprule
\multicolumn{1}{l|}{\textbf{Dataset}} & \multicolumn{1}{c}{\textbf{First set sentence}} & \multicolumn{1}{|l}{\textbf{Second set sentence}} \\
\midrule
\midrule
BUCC & \multicolumn{1}{p{5cm}|}{Morales remporte l’élection présidentielle de 2005 à la majorité absolue.} & \multicolumn{1}{p{5cm}}{Morales went on to win the 2005 presidential election with an absolute majority.} \\
\midrule
Tatoeba & \multicolumn{1}{p{5cm}|}{Chi le ha detto che Tom l'ha fatto?} & \multicolumn{1}{p{5cm}}{Who told you that Tom did that?} \\
    \bottomrule
    \end{tabular}}
    \caption{Bitext mining examples}
    \label{tab:bitext_mining_examples}
\end{table*}

\begin{table*}[!t]
    \centering
    \tiny
    \resizebox{\textwidth}{!}{\begin{tabular}{l|l|l|c}
    \toprule
\multicolumn{1}{l|}{\textbf{Dataset}} & \multicolumn{1}{c}{\textbf{Human Summary}} & \multicolumn{1}{|l}{\textbf{Machine Summary}} & \multicolumn{1}{|l}{\textbf{Relevance}} \\
\midrule
\midrule
SummEval & \multicolumn{1}{p{5cm}|}{V. Stiviano must pay back \$2.6 million in gifts from Donald Sterling. Sterling's wife claimed the ex-Clippers used the couple's money for the gifts. The items included a Ferrari, two Bentleys and a Range Rover.} & \multicolumn{1}{p{5cm}|}{donald sterling , nba team last year . sterling 's wife sued for \$ 2.6 million in gifts . sterling says he is the former female companion who has lost the . sterling has ordered v. stiviano to pay back \$ 2.6 m in gifts after his wife sued . sterling also includes a \$ 391 easter bunny costume , \$ 299 and a \$ 299 .} & 1.7 \\
    \bottomrule
    \end{tabular}}
    \caption{Summarization example}
    \label{tab:summarization_examples}
\end{table*}

\section{Correlations}
\label{sec:corr}

\begin{figure*}
    \centering
    \subfloat[\centering Model correlation based on all results]{{\includegraphics[width=0.45\textwidth]{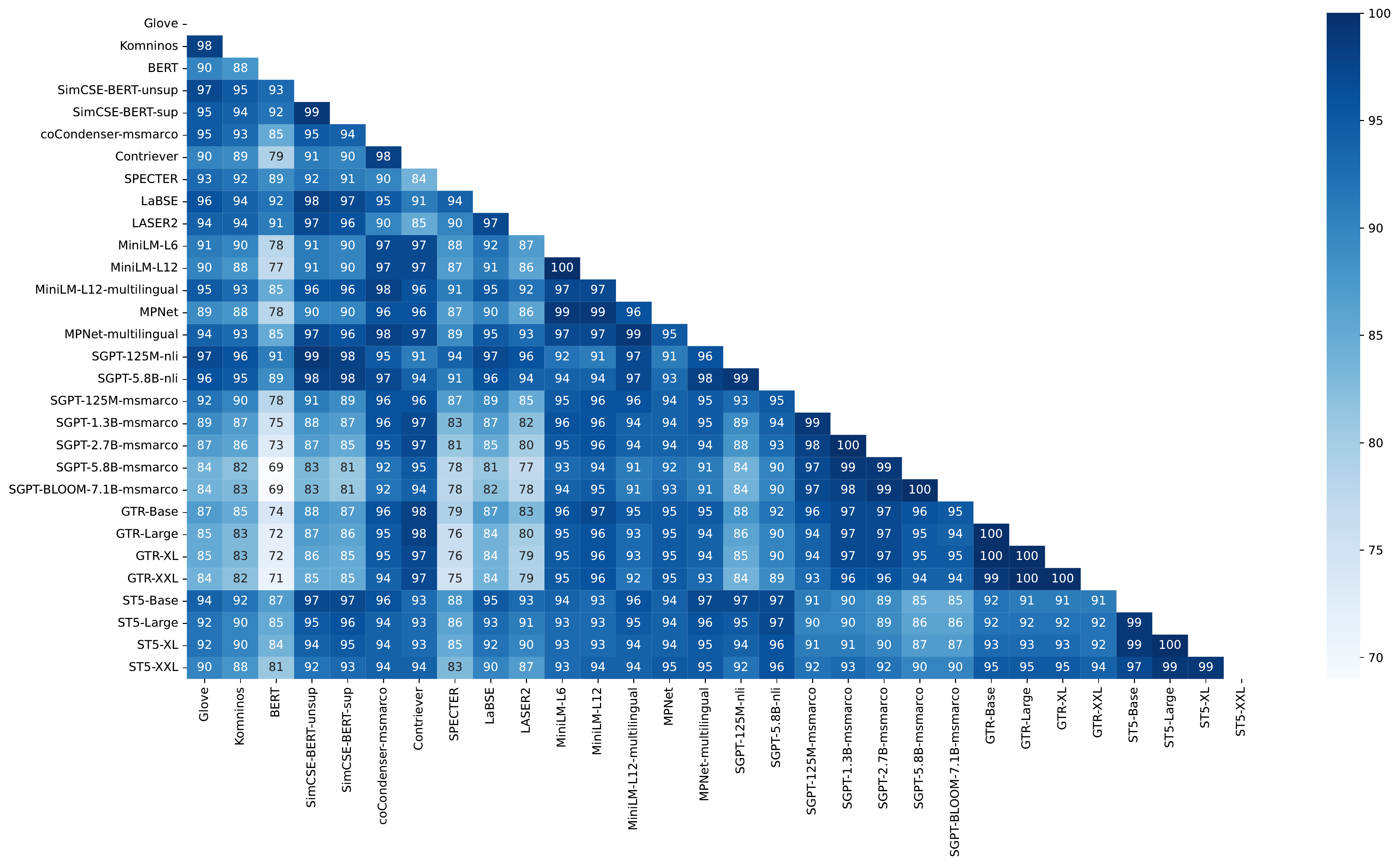}}}
    \qquad
    \subfloat[\centering Task correlation based on average task results]{{\includegraphics[width=0.45\textwidth]{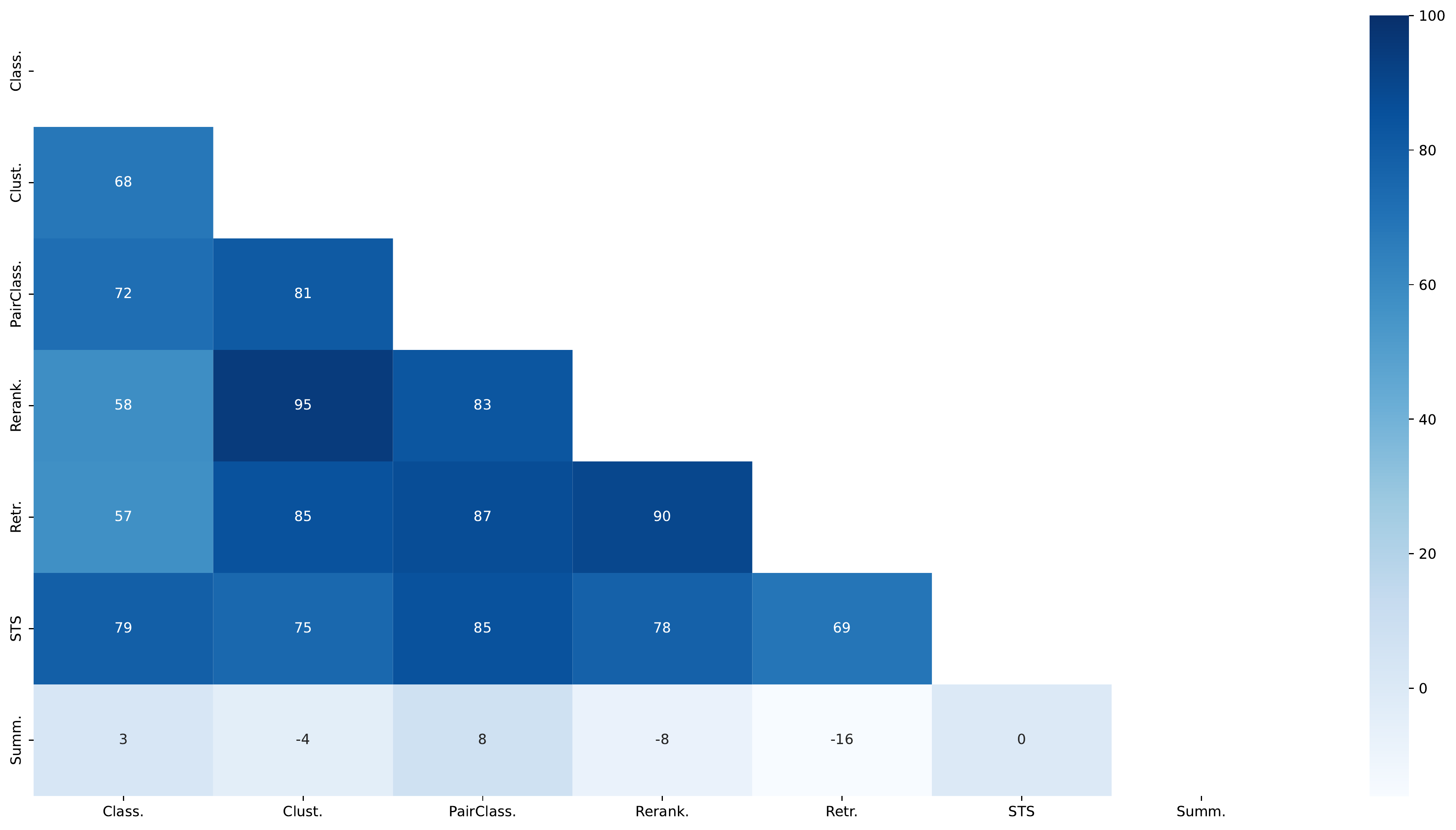}}}
    \caption{Pearson correlations across model and task results. \textbf{Left:} Size variants of the same architecture show high correlations. \textbf{Right:} Performance on clustering and reranking correlates strongest, while summarization and classification show weaker correlation with other tasks.}
    \label{fig:corrs}
\end{figure*}

Figure \ref{fig:corrs} provides correlation heatmaps for model performance and MTEB tasks.

\section{Models}
\label{sec:appmodels}

Table \ref{tab:ckpts} provides publicly available model checkpoints used for MTEB evaluation.

\begin{table*}[!t]
    \centering
    \tiny
    \resizebox{\textwidth}{!}{\begin{tabular}{l|l}
    \toprule
\multicolumn{1}{l|}{\textbf{Model}} & \multicolumn{1}{c}{\textbf{Public Checkpoint}} \\
\midrule
\midrule
Glove & \url{https://huggingface.co/sentence-transformers/average_word_embeddings_glove.6B.300d} \\
Komninos & \url{https://huggingface.co/sentence-transformers/average_word_embeddings_komninos} \\
BERT & \url{https://huggingface.co/bert-base-uncased} \\
SimCSE-BERT-unsup & \url{https://huggingface.co/princeton-nlp/unsup-simcse-bert-base-uncased} \\
SimCSE-BERT-sup & \url{https://huggingface.co/princeton-nlp/sup-simcse-bert-base-uncased} \\
coCondenser-msmarco & \url{https://huggingface.co/sentence-transformers/msmarco-bert-co-condensor} \\
Contriever & \url{https://huggingface.co/nthakur/contriever-base-msmarco} \\
SPECTER & \url{https://huggingface.co/sentence-transformers/allenai-specter} \\
LaBSE & \url{https://huggingface.co/sentence-transformers/LaBSE} \\
LASER2 & \url{https://github.com/facebookresearch/LASER} \\
MiniLM-L6 & \url{https://huggingface.co/sentence-transformers/all-MiniLM-L6-v2} \\
MiniLM-L12 & \url{https://huggingface.co/sentence-transformers/all-MiniLM-L12-v2} \\
MiniLM-L12-multilingual & \url{https://huggingface.co/sentence-transformers/paraphrase-multilingual-MiniLM-L12-v2} \\
MPNet & \url{https://huggingface.co/sentence-transformers/all-mpnet-base-v2} \\
MPNet-multilingual & \url{https://huggingface.co/sentence-transformers/paraphrase-multilingual-mpnet-base-v2} \\
MiniLM-L12-multilingual & \url{https://huggingface.co/sentence-transformers/paraphrase-multilingual-MiniLM-L12-v2} \\
SGPT-125M-nli & \url{https://huggingface.co/Muennighoff/SGPT-125M-weightedmean-nli-bitfit} \\
SGPT-5.8B-nli & \url{https://huggingface.co/Muennighoff/SGPT-5.8B-weightedmean-nli-bitfit} \\
SGPT-125M-msmarco & \url{https://huggingface.co/Muennighoff/SGPT-125M-weightedmean-msmarco-specb-bitfit} \\
SGPT-1.3B-msmarco & \url{https://huggingface.co/Muennighoff/SGPT-1.3B-weightedmean-msmarco-specb-bitfit} \\
SGPT-2.7B-msmarco & \url{https://huggingface.co/Muennighoff/SGPT-2.7B-weightedmean-msmarco-specb-bitfit} \\
SGPT-5.8B-msmarco & \url{https://huggingface.co/Muennighoff/SGPT-5.8B-weightedmean-msmarco-specb-bitfit} \\
SGPT-BLOOM-7.1B-msmarco & \url{https://huggingface.co/bigscience/sgpt-bloom-7b1-msmarco} \\
SGPT-BLOOM-1.7B-nli & \url{https://huggingface.co/bigscience-data/sgpt-bloom-1b7-nli} \\
GTR-Base & \url{https://huggingface.co/sentence-transformers/gtr-t5-base} \\
GTR-Large & \url{https://huggingface.co/sentence-transformers/gtr-t5-large} \\
GTR-XL & \url{https://huggingface.co/sentence-transformers/gtr-t5-xl} \\
GTR-XXL & \url{https://huggingface.co/sentence-transformers/gtr-t5-xxl} \\
ST5-Base & \url{https://huggingface.co/sentence-transformers/sentence-t5-base} \\
ST5-Large & \url{https://huggingface.co/sentence-transformers/sentence-t5-large} \\
ST5-XL & \url{https://huggingface.co/sentence-transformers/sentence-t5-xl} \\
ST5-XXL & \url{https://huggingface.co/sentence-transformers/sentence-t5-xxl} \\
    \bottomrule
    \end{tabular}}
    \caption{Publicly available model links used for evaluation}
    \label{tab:ckpts}
\end{table*}

\section{Additional results}
\label{sec:addresults}

Tables \ref{tab:addresults} until the end provide results on individual datasets of MTEB. The results are additionally available in json format on the Hugging Face Hub\footnote{\url{https://huggingface.co/datasets/mteb/results}} and can be inspected on the leaderboard\footnote{\url{https://huggingface.co/spaces/mteb/leaderboard}}.

\newpage

\begin{landscape}
\begin{table*}[!t]
    \begin{adjustwidth}{-11cm}{}
    \centering
    \tiny
    \resizebox{1.84\textwidth}{!}{
    
    % To fit it into A4 (for EACL) remove adjustwidth, landscape, resize box above & uncomment the below
    %\resizebox{\textwidth}{!}{

    \begin{tabular}{l|ccccccccccccccccccccccccccccccccccc}
    \toprule
Dataset & Glove & Komninos & BERT & SimCSE- & SimCSE- & coCondenser- & Contr- & SPECTER & LaBSE & LASER2 & MiniLM- & MiniLM- & MiniLM- & MPNet & MPNet- & OpenAI & SGPT-125M- & SGPT-5.8B- & SGPT-125M- & SGPT-1.3B- & SGPT-2.7B- & SGPT-5.8B- & SGPT- & GTR- & GTR- & GTR- & GTR- & ST5- & ST5- & ST5- & ST5- \\
 & & & & BERT- & BERT- & msmarco & iever & & & & L6 & L12- & L12- & & multilingual & Ada & nli & nli & msmarco & msmarco & msmarco & msmarco & BLOOM-7.1B- & Base & Large & XL & XXL & Base & Large & XL & XXL \\
 & & & & unsup & sup & & & & & & & & multilingual & & & Similarity & & & & & & & msmarco & & & & & & & & \\
\midrule
\midrule
AmazonCounterfactualClassification & 56.91 & 60.54 & 74.25 & 67.09 & 75.75 & 64.06 & 72.19 & 58.70 & 75.93 & 76.84 & 64.15 & 65.28 & 71.57 & 65.27 & 75.81 & 76.40 & 65.88 & 74.07 & 61.24 & 65.21 & 67.57 & 69.22 & 68.06 & 69.33 & 70.03 & 68.60 & 67.30 & 75.82 & 75.51 & 76.01 & 77.07 \\
AmazonPolarityClassification & 60.32 & 59.59 & 71.33 & 74.48 & 82.47 & 66.88 & 68.63 & 57.77 & 68.95 & 61.01 & 62.58 & 62.98 & 69.21 & 67.13 & 76.41 & 92.83 & 74.94 & 82.31 & 65.40 & 73.21 & 71.44 & 71.26 & 68.97 & 67.82 & 73.92 & 74.58 & 75.05 & 85.12 & 92.87 & 93.17 & 92.79 \\
AmazonReviewsClassification & 29.67 & 31.01 & 33.56 & 33.85 & 39.60 & 34.85 & 37.42 & 26.26 & 35.80 & 28.71 & 31.79 & 30.79 & 35.11 & 31.92 & 38.51 & 47.45 & 35.10 & 41.58 & 31.17 & 34.96 & 35.75 & 39.19 & 33.86 & 38.48 & 37.21 & 38.20 & 37.30 & 44.94 & 47.12 & 48.18 & 48.93 \\
Banking77Classification & 67.69 & 67.05 & 63.41 & 73.55 & 75.76 & 82.35 & 80.02 & 66.66 & 69.85 & 57.76 & 79.75 & 80.40 & 79.77 & 81.86 & 81.07 & 68.04 & 74.68 & 81.74 & 77.70 & 82.06 & 83.22 & 84.49 & 84.33 & 79.26 & 81.21 & 82.22 & 82.32 & 76.48 & 78.46 & 80.88 & 82.31 \\
EmotionClassification & 36.93 & 33.18 & 35.28 & 42.22 & 44.81 & 41.91 & 44.77 & 24.82 & 37.22 & 24.83 & 38.43 & 41.17 & 42.37 & 39.73 & 45.84 & 50.32 & 42.23 & 49.92 & 39.08 & 46.39 & 49.21 & 49.66 & 44.87 & 42.20 & 46.32 & 45.55 & 43.19 & 51.36 & 51.73 & 51.95 & 48.57 \\
ImdbClassification & 62.57 & 63.98 & 65.35 & 69.63 & 73.53 & 60.17 & 67.04 & 56.35 & 62.04 & 57.58 & 60.66 & 59.76 & 60.46 & 70.72 & 64.57 & 89.38 & 62.90 & 74.33 & 58.67 & 64.05 & 63.53 & 66.64 & 61.77 & 65.99 & 70.86 & 68.15 & 70.8 & 77.34 & 87.01 & 87.54 & 90.23 \\
MassiveIntentClassification & 56.19 & 57.21 & 59.88 & 59.84 & 65.95 & 70.40 & 67.78 & 51.73 & 61.46 & 47.91 & 67.40 & 67.15 & 66.84 & 69.57 & 69.32 & 65.17 & 58.08 & 70.0 & 61.41 & 68.65 & 69.01 & 70.39 & 69.67 & 67.05 & 70.06 & 70.23 & 70.61 & 69.74 & 71.78 & 72.09 & 73.44 \\
MassiveScenarioClassification & 66.03 & 66.11 & 64.28 & 66.25 & 70.78 & 73.73 & 76.00 & 58.58 & 66.41 & 55.92 & 75.76 & 74.58 & 71.51 & 76.01 & 75.35 & 67.67 & 66.34 & 75.03 & 69.74 & 76.04 & 75.90 & 76.28 & 75.34 & 75.40 & 75.49 & 75.94 & 77.77 & 72.32 & 73.16 & 73.26 & 74.82 \\
MTOPDomainClassification & 79.11 & 78.57 & 82.63 & 81.71 & 84.29 & 91.34 & 93.18 & 74.53 & 86.06 & 75.36 & 91.56 & 91.90 & 87.06 & 92.08 & 89.24 & 89.89 & 81.52 & 89.64 & 86.96 & 92.08 & 92.56 & 93.47 & 93.68 & 92.42 & 94.01 & 93.60 & 93.84 & 90.34 & 90.99 & 90.73 & 92.49 \\
MTOPIntentClassification & 55.85 & 57.07 & 68.14 & 59.23 & 63.14 & 71.07 & 69.31 & 50.05 & 63.03 & 49.47 & 62.18 & 62.84 & 65.52 & 70.21 & 68.69 & 64.80 & 58.24 & 70.68 & 62.25 & 71.19 & 71.85 & 72.42 & 71.34 & 62.44 & 63.86 & 65.93 & 67.71 & 63.32 & 64.98 & 68.15 & 68.33 \\
ToxicConversationsClassification & 65.40 & 67.76 & 70.0 & 68.82 & 72.04 & 64.01 & 67.77 & 57.44 & 66.90 & 54.05 & 66.99 & 67.47 & 66.07 & 60.86 & 71.02 & 70.00 & 62.79 & 69.93 & 62.66 & 68.73 & 68.84 & 67.71 & 66.55 & 66.60 & 68.65 & 67.56 & 68.48 & 68.20 & 71.73 & 70.95 & 70.04 \\
TweetSentimentExtractionClassification & 50.80 & 49.68 & 51.81 & 53.36 & 59.73 & 55.74 & 56.10 & 45.52 & 58.82 & 48.73 & 55.41 & 54.25 & 56.12 & 55.46 & 59.03 & 63.35 & 54.82 & 62.44 & 52.41 & 55.67 & 56.69 & 56.85 & 55.85 & 56.02 & 54.09 & 54.77 & 54.54 & 62.71 & 62.33 & 61.21 & 62.01 \\
\midrule
ArxivClusteringP2P & 32.56 & 34.73 & 35.19 & 32.61 & 35.18 & 36.94 & 42.61 & 44.75 & 32.13 & 17.77 & 46.55 & 46.07 & 38.33 & 48.38 & 37.78 & 41.49 & 34.74 & 40.55 & 39.71 & 43.38 & 44.72 & 45.59 & 44.59 & 35.49 & 37.50 & 37.90 & 37.90 & 39.28 & 41.62 & 41.62 & 42.89 \\
ArxivClusteringS2S & 23.14 & 26.01 & 27.51 & 24.68 & 27.54 & 29.03 & 32.32 & 35.27 & 22.05 & 12.39 & 37.86 & 37.50 & 31.55 & 39.72 & 31.68 & 28.47 & 24.68 & 32.49 & 28.24 & 33.71 & 35.08 & 38.86 & 38.03 & 27.18 & 30.55 & 30.45 & 32.39 & 27.26 & 29.44 & 31.17 & 33.47 \\
BiorxivClusteringP2P & 29.27 & 29.76 & 30.12 & 24.90 & 30.15 & 32.35 & 34.97 & 39.52 & 29.84 & 12.40 & 38.48 & 36.99 & 33.49 & 39.62 & 33.09 & 36.86 & 28.93 & 33.59 & 33.63 & 35.06 & 34.41 & 36.55 & 36.03 & 27.66 & 29.59 & 30.52 & 30.48 & 33.99 & 35.99 & 36.43 & 36.53 \\
BiorxivClusteringS2S & 19.18 & 20.71 & 24.77 & 19.55 & 24.67 & 28.16 & 29.08 & 34.53 & 20.57 & 8.83 & 33.17 & 33.21 & 29.44 & 35.02 & 29.60 & 27.55 & 23.08 & 29.13 & 27.04 & 30.71 & 30.53 & 33.70 & 32.48 & 23.25 & 25.72 & 26.06 & 27.50 & 22.92 & 24.02 & 26.47 & 28.66 \\
MedrxivClusteringP2P & 26.12 & 26.65 & 26.09 & 23.60 & 26.25 & 30.23 & 31.19 & 35.04 & 30.13 & 17.91 & 34.41 & 34.25 & 31.52 & 35.58 & 31.96 & 31.09 & 28.30 & 30.33 & 31.37 & 32.08 & 31.35 & 31.51 & 31.05 & 27.57 & 28.72 & 28.69 & 29.12 & 33.20 & 32.40 & 32.30 & 32.09 \\
MedrxivClusteringS2S & 20.38 & 21.50 & 23.60 & 21.97 & 24.12 & 27.01 & 27.27 & 31.66 & 24.82 & 16.63 & 32.29 & 32.24 & 30.87 & 32.87 & 31.70 & 26.50 & 24.93 & 28.02 & 26.87 & 29.45 & 28.77 & 28.76 & 29.26 & 25.13 & 27.39 & 26.69 & 27.56 & 26.13 & 26.33 & 26.93 & 26.82 \\
RedditClustering & 28.46 & 28.84 & 27.24 & 32.18 & 40.23 & 48.04 & 54.89 & 24.13 & 28.79 & 9.96 & 50.67 & 51.18 & 42.02 & 54.82 & 45.24 & 42.47 & 33.76 & 42.17 & 40.23 & 48.23 & 46.47 & 40.45 & 35.53 & 56.13 & 61.69 & 61.34 & 64.13 & 52.93 & 54.53 & 57.03 & 58.99 \\
RedditClusteringP2P & 35.82 & 7.37 & 43.32 & 45.14 & 47.74 & 53.53 & 57.58 & 35.06 & 49.14 & 26.42 & 54.15 & 54.80 & 50.73 & 56.77 & 51.31 & 58.10 & 41.01 & 48.02 & 49.09 & 53.18 & 54.17 & 55.75 & 54.52 & 58.53 & 61.67 & 61.11 & 62.84 & 59.67 & 62.50 & 62.34 & 64.46 \\
StackExchangeClustering & 35.80 & 39.04 & 43.58 & 43.07 & 47.55 & 59.54 & 63.15 & 39.01 & 35.43 & 15.79 & 53.36 & 53.05 & 49.60 & 53.80 & 52.98 & 53.52 & 44.59 & 54.13 & 52.74 & 60.86 & 59.19 & 59.21 & 55.13 & 64.21 & 69.93 & 69.95 & 71.43 & 63.13 & 65.11 & 67.13 & 70.78 \\
StackExchangeClusteringP2P & 28.51 & 30.23 & 26.55 & 28.50 & 29.45 & 30.48 & 32.25 & 31.46 & 28.83 & 18.63 & 38.00 & 33.13 & 31.69 & 34.28 & 32.94 & 30.43 & 28.23 & 31.12 & 32.66 & 32.36 & 32.57 & 33.95 & 34.31 & 33.01 & 33.21 & 32.73 & 32.85 & 35.68 & 36.86 & 34.79 & 35.25 \\
TwentyNewsgroupsClustering & 25.83 & 27.42 & 23.35 & 23.21 & 34.86 & 38.68 & 46.82 & 24.22 & 23.28 & 11.38 & 46.86 & 47.47 & 39.28 & 49.74 & 44.10 & 36.26 & 28.24 & 37.20 & 32.13 & 40.06 & 40.89 & 39.46 & 37.28 & 46.72 & 51.64 & 51.15 & 50.44 & 48.10 & 49.33 & 49.53 & 50.93 \\
\midrule
SprintDuplicateQuestions & 86.96 & 85.55 & 36.81 & 69.41 & 69.39 & 96.09 & 95.55 & 71.63 & 89.26 & 65.54 & 94.55 & 92.45 & 89.46 & 90.15 & 90.55 & 77.85 & 77.73 & 80.54 & 89.89 & 92.58 & 93.47 & 93.84 & 94.93 & 94.55 & 95.05 & 95.45 & 95.68 & 91.23 & 89.01 & 91.44 & 88.89 \\
TwitterSemEval2015 & 48.45 & 53.85 & 55.90 & 60.21 & 67.75 & 65.95 & 66.85 & 43.25 & 62.78 & 59.57 & 67.86 & 70.02 & 62.06 & 73.85 & 66.75 & 69.04 & 57.09 & 66.00 & 54.75 & 62.37 & 63.68 & 66.87 & 65.31 & 72.23 & 76.03 & 77.81 & 77.54 & 78.25 & 79.75 & 80.89 & 80.28 \\
TwitterURLCorpus & 77.35 & 79.41 & 76.29 & 81.37 & 83.89 & 83.17 & 85.21 & 69.22 & 84.58 & 81.47 & 84.70 & 84.77 & 83.83 & 85.11 & 85.14 & 83.69 & 80.51 & 84.54 & 81.06 & 83.79 & 84.80 & 85.29 & 85.46 & 84.77 & 84.89 & 85.14 & 85.13 & 86.05 & 86.14 & 85.86 & 86.01 \\
\midrule
AskUbuntuDupQuestions & 49.57 & 50.88 & 45.84 & 51.57 & 51.80 & 58.99 & 56.69 & 50.07 & 52.75 & 48.99 & 63.48 & 64.06 & 60.49 & 65.85 & 60.16 & 53.49 & 52.63 & 55.90 & 55.84 & 58.13 & 59.63 & 61.63 & 59.97 & 60.86 & 61.64 & 63.08 & 63.23 & 59.73 & 61.51 & 62.86 & 66.16 \\
MindSmallReranking & 27.01 & 28.92 & 28.37 & 28.62 & 29.30 & 27.13 & 31.58 & 24.80 & 29.81 & 24.79 & 30.80 & 31.02 & 30.37 & 30.97 & 30.15 & 30.71 & 29.27 & 31.11 & 30.40 & 31.34 & 31.72 & 32.29 & 31.79 & 31.33 & 31.84 & 31.50 & 31.93 & 30.20 & 30.27 & 29.77 & 30.60 \\
SciDocsRR & 62.56 & 63.55 & 64.94 & 66.33 & 70.14 & 72.78 & 76.51 & 81.31 & 68.72 & 54.99 & 87.12 & 87.20 & 77.78 & 88.65 & 78.09 & 71.04 & 68.36 & 77.54 & 71.34 & 77.21 & 77.72 & 80.79 & 79.77 & 73.71 & 76.39 & 76.49 & 77.96 & 73.96 & 74.88 & 75.16 & 76.09 \\
StackOverflowDupQuestions & 34.03 & 35.65 & 34.62 & 39.35 & 38.90 & 48.48 & 47.78 & 36.22 & 42.42 & 36.98 & 50.76 & 51.47 & 45.85 & 51.98 & 46.79 & 40.85 & 39.97 & 44.77 & 44.74 & 49.32 & 49.61 & 51.53 & 51.07 & 51.01 & 51.58 & 52.79 & 53.50 & 48.46 & 49.34 & 51.05 & 52.85 \\
\midrule
ArguAna & 36.30 & 30.96 & 28.29 & 38.34 & 38.33 & 45.15 & 48.32 & 32.67 & 34.18 & 12.86 & 50.17 & 47.13 & 44.88 & 46.52 & 48.91 & 39.65 & 31.04 & 35.07 & 45.42 & 49.68 & 50.49 & 51.38 & 47.28 & 50.83 & 52.09 & 52.81 & 53.77 & 44.85 & 39.27 & 39.40 & 39.85 \\
ClimateFEVER & 14.44 & 14.87 & 5.41 & 11.80 & 11.98 & 16.96 & 24.79 & 6.86 & 3.83 & 0.36 & 20.27 & 21.57 & 18.49 & 21.97 & 15.27 & 2.83 & 11.01 & 17.57 & 21.86 & 26.6 & 27.11 & 30.46 & 29.39 & 24.88 & 26.90 & 27.01 & 27.21 & 10.37 & 11.36 & 10.61 & 14.63 \\
CQADupstackRetrieval & 15.47 & 16.79 & 5.51 & 13.22 & 14.50 & 27.72 & 33.67 & 14.60 & 18.75 & 4.12 & 41.32 & 42.53 & 30.71 & 44.96 & 31.32 & 10.17 & 20.29 & 29.98 & 27.25 & 33.33 & 36.53 & 39.40 & 39.62 & 34.55 & 36.62 & 37.35 & 38.56 & 35.23 & 38.96 & 40.78 & 44.65 \\
DBPedia & 18.29 & 15.88 & 4.13 & 15.04 & 19.73 & 27.86 & 38.10 & 4.14 & 15.57 & 1.53 & 32.33 & 33.36 & 22.63 & 32.09 & 26.22 & 3.48 & 10.87 & 26.10 & 22.72 & 31.51 & 34.70 & 39.87 & 39.03 & 35.24 & 39.55 & 39.74 & 41.28 & 27.77 & 31.55 & 33.65 & 39.19 \\
FEVER & 14.99 & 15.56 & 3.30 & 21.05 & 20.41 & 45.68 & 59.29 & 5.45 & 12.17 & 0.77 & 51.93 & 55.91 & 52.66 & 50.86 & 56.76 & 4.45 & 18.40 & 38.64 & 60.45 & 68.12 & 72.73 & 78.24 & 73.97 & 68.93 & 72.66 & 72.18 & 74.08 & 26.16 & 36.21 & 36.12 & 51.20 \\
FiQA2018 & 10.09 & 10.49 & 2.19 & 9.84 & 10.41 & 15.62 & 27.42 & 5.64 & 7.00 & 1.73 & 36.87 & 37.27 & 20.33 & 49.96 & 22.96 & 7.54 & 8.94 & 18.59 & 21.12 & 29.99 & 33.29 & 37.20 & 35.84 & 35.15 & 42.79 & 44.19 & 46.78 & 34.83 & 43.55 & 44.71 & 46.68 \\
HotpotQA & 19.18 & 20.77 & 8.26 & 19.75 & 22.89 & 35.61 & 56.81 & 5.46 & 18.75 & 5.50 & 46.51 & 44.59 & 30.01 & 39.29 & 37.03 & 12.6 & 17.73 & 33.99 & 40.88 & 49.93 & 52.84 & 59.26 & 57.26 & 54.93 & 57.85 & 58.91 & 59.67 & 33.20 & 33.95 & 37.17 & 42.14 \\
MSMARCO & 9.60 & 9.75 & 1.91 & 9.35 & 11.00 & 29.57 & 36.77 & 5.58 & 7.60 & 1.09 & 36.54 & 39.03 & 23.72 & 39.75 & 26.60 & 10.53 & 6.27 & 15.83 & 27.98 & 36.05 & 38.83 & 39.91 & 41.12 & 41.16 & 42.73 & 43.52 & 44.05 & 20.71 & 23.96 & 25.17 & 27.68 \\
NFCorpus & 13.87 & 11.79 & 4.30 & 9.88 & 12.42 & 22.29 & 31.31 & 0.84 & 16.54 & 2.44 & 31.59 & 32.25 & 23.45 & 33.29 & 25.49 & 20.59 & 11.80 & 28.26 & 22.79 & 32.08 & 33.89 & 36.21 & 35.78 & 30.22 & 32.63 & 33.34 & 34.18 & 28.64 & 31.10 & 33.18 & 35.08 \\
NQ & 12.87 & 12.75 & 2.61 & 11.69 & 16.08 & 29.85 & 41.83 & 5.99 & 8.42 & 0.64 & 43.87 & 46.47 & 29.80 & 50.45 & 33.60 & 2.02 & 7.63 & 24.63 & 29.73 & 42.94 & 46.70 & 52.41 & 53.15 & 50.47 & 55.09 & 56.16 & 57.24 & 36.32 & 42.02 & 46.29 & 52.87 \\
QuoraRetrieval & 71.32 & 71.58 & 61.03 & 78.03 & 79.62 & 86.51 & 86.72 & 64.65 & 77.03 & 71.14 & 87.56 & 87.75 & 86.55 & 87.46 & 86.41 & 82.18 & 78.96 & 84.68 & 72.98 & 85.28 & 85.60 & 84.58 & 74.71 & 87.98 & 88.47 & 88.91 & 89.09 & 85.49 & 85.73 & 85.85 & 85.96 \\
SCIDOCS & 8.04 & 8.47 & 2.81 & 5.50 & 7.53 & 10.13 & 17.12 & 0.00 & 5.63 & 0.78 & 21.64 & 21.82 & 0.03 & 23.77 & 13.96 & 6.28 & 7.13 & 13.55 & 12.21 & 16.18 & 16.57 & 19.87 & 18.62 & 14.00 & 15.51 & 15.71 & 15.88 & 14.16 & 15.38 & 15.97 & 17.17 \\
SciFact & 29.58 & 29.53 & 13.34 & 25.72 & 29.59 & 52.31 & 65.51 & 47.88 & 38.20 & 4.04 & 64.51 & 62.64 & 48.37 & 65.57 & 50.30 & 45.46 & 31.79 & 46.66 & 56.90 & 68.29 & 70.17 & 74.70 & 72.11 & 59.74 & 63.42 & 64.20 & 66.77 & 45.76 & 49.91 & 50.91 & 55.38 \\
Touche2020 & 13.99 & 13.17 & 0.97 & 8.90 & 9.89 & 8.57 & 15.79 & 8.46 & 4.88 & 1.06 & 16.90 & 17.22 & 16.06 & 19.93 & 17.40 & 3.1 & 12.27 & 16.18 & 22.97 & 24.45 & 23.44 & 25.43 & 23.98 & 25.89 & 28.29 & 25.26 & 26.76 & 20.30 & 21.63 & 22.51 & 21.65 \\
TRECCOVID & 36.22 & 35.92 & 14.74 & 26.2 & 22.93 & 40.54 & 44.77 & 29.91 & 16.34 & 10.97 & 47.25 & 50.82 & 39.12 & 51.33 & 37.87 & 24.56 & 39.31 & 55.35 & 70.30 & 72.98 & 75.17 & 84.88 & 81.37 & 56.05 & 56.68 & 60.09 & 51.90 & 40.70 & 46.11 & 54.77 & 59.48 \\
\midrule
BIOSSES & 44.93 & 50.25 & 54.70 & 72.31 & 68.38 & 77.32 & 83.32 & 64.95 & 78.70 & 62.01 & 81.64 & 83.57 & 74.18 & 80.43 & 76.27 & 78.04 & 70.93 & 79.50 & 75.21 & 83.02 & 84.84 & 86.25 & 85.31 & 79.00 & 84.86 & 78.94 & 81.91 & 75.89 & 78.93 & 73.12 & 80.43 \\
SICK-R & 55.43 & 55.49 & 58.65 & 72.24 & 80.77 & 72.00 & 70.20 & 56.39 & 69.99 & 62.86 & 77.58 & 79.32 & 79.61 & 80.59 & 79.62 & 77.48 & 74.57 & 79.59 & 65.93 & 67.23 & 68.20 & 69.63 & 69.82 & 71.45 & 73.39 & 73.63 & 74.29 & 80.18 & 80.34 & 79.98 & 80.47 \\
STS12 & 54.64 & 53.51 & 30.87 & 66.05 & 75.30 & 68.19 & 64.34 & 62.49 & 65.08 & 62.60 & 72.37 & 73.08 & 76.02 & 72.63 & 77.90 & 72.30 & 69.17 & 74.29 & 66.53 & 66.59 & 66.99 & 67.50 & 69.66 & 68.59 & 70.33 & 69.11 & 70.12 & 78.05 & 79.11 & 79.02 & 78.85 \\
STS13 & 69.16 & 70.80 & 59.89 & 81.49 & 84.67 & 80.40 & 80.03 & 58.70 & 67.98 & 59.62 & 80.60 & 82.13 & 80.70 & 83.48 & 85.11 & 81.49 & 77.23 & 85.35 & 76.17 & 77.33 & 77.58 & 79.16 & 79.67 & 79.09 & 82.19 & 81.82 & 82.72 & 85.85 & 87.33 & 88.80 & 88.94 \\
STS14 & 60.81 & 63.56 & 47.73 & 73.61 & 80.19 & 74.02 & 74.51 & 54.87 & 64.03 & 57.03 & 75.59 & 76.73 & 78.85 & 78.00 & 80.81 & 74.74 & 70.99 & 79.21 & 69.05 & 71.83 & 72.78 & 74.46 & 74.61 & 74.64 & 77.16 & 77.07 & 78.24 & 82.19 & 83.17 & 84.33 & 84.86 \\
STS15 & 72.31 & 74.08 & 60.29 & 79.72 & 85.40 & 82.57 & 83.30 & 62.54 & 76.59 & 71.57 & 85.39 & 85.58 & 85.84 & 85.66 & 87.48 & 84.28 & 79.74 & 85.52 & 79.24 & 80.66 & 82.62 & 84.47 & 83.81 & 84.85 & 86.31 & 86.01 & 86.26 & 87.46 & 88.28 & 88.89 & 89.32 \\
STS16 & 65.34 & 64.60 & 63.73 & 78.12 & 80.82 & 79.78 & 79.67 & 64.27 & 72.98 & 70.75 & 78.99 & 80.23 & 81.05 & 80.03 & 83.20 & 82.06 & 77.93 & 82.54 & 76.07 & 78.91 & 80.10 & 80.96 & 80.40 & 81.57 & 81.85 & 82.23 & 81.61 & 84.03 & 84.36 & 85.31 & 84.67 \\
STS17 & 77.95 & 76.91 & 64.10 & 83.58 & 89.44 & 85.94 & 86.32 & 69.63 & 79.45 & 76.73 & 87.59 & 88.63 & 86.87 & 90.60 & 86.99 & 87.08 & 87.33 & 90.44 & 84.95 & 86.99 & 87.25 & 87.78 & 87.07 & 85.80 & 83.93 & 84.90 & 85.18 & 89.57 & 88.99 & 88.91 & 89.46 \\
STS22 & 56.35 & 53.89 & 56.37 & 59.65 & 61.96 & 67.54 & 64.64 & 55.06 & 60.97 & 39.75 & 67.21 & 65.67 & 61.72 & 67.95 & 63.06 & 64.71 & 59.64 & 63.20 & 65.66 & 67.30 & 68.75 & 69.35 & 66.13 & 66.17 & 64.30 & 66.61 & 65.76 & 62.66 & 62.39 & 64.32 & 65.33 \\
STSBenchmark & 61.54 & 61.55 & 47.29 & 76.52 & 84.25 & 76.97 & 78.81 & 61.26 & 72.25 & 69.77 & 82.03 & 83.09 & 84.42 & 83.42 & 86.82 & 83.78 & 79.54 & 85.67 & 75.34 & 77.59 & 79.21 & 81.39 & 80.90 & 79.58 & 77.60 & 77.65 & 77.73 & 85.52 & 85.36 & 83.93 & 84.01 \\
\midrule
SummEval & 28.87 & 30.49 & 29.82 & 31.15 & 23.31 & 29.50 & 30.36 & 27.66 & 31.05 & 26.8 & 30.81 & 27.9 & 30.67 & 27.49 & 31.57 & 26.94 & 30.26 & 30.38 & 28.90 & 25.44 & 27.87 & 24.75 & 24.99 & 29.67 & 29.50 & 30.21 & 30.64 & 31.39 & 29.64 & 29.91 & 30.08 \\
\midrule
\midrule
Average & 41.97 & 42.06 & 38.33 & 45.45 & 48.72 & 52.35 & 56.00 & 40.28 & 45.21 & 34.95 & 56.26 & 56.53 & 52.44 & 57.78 & 54.71 & 49.52 & 45.97 & 53.74 & 51.23 & 56.11 & 57.12 & 58.81 & 57.44 & 56.19 & 58.28 & 58.42 & 58.97 & 55.27 & 57.06 & 57.87 & 59.51 \\
    \bottomrule
    \end{tabular}
    }
    \end{adjustwidth}
    \begin{adjustwidth}{-5.2cm}{}
    \caption{All English results. The main score for each task is reported as described in Section \ref{sec:taskeval}.}
    \label{tab:addresults}
    \end{adjustwidth}
\end{table*}

\end{landscape}

\newpage

\begin{table*}[t!]
    \centering
    \resizebox{0.52\textwidth}{!}{\begin{tabular}{lc|cccccc}
    \toprule
Dataset & Language & LASER2 & LaBSE & MiniLM-L12-multilingual & MPNet-multilingual & SGPT-BLOOM-7.1B-msmarco \\
\midrule
\midrule
BUCC & de-en & 99.21 & 99.35 & 97.11 & 98.59 & 54.00 \\
BUCC & fr-en & 98.39 & 98.72 & 94.99 & 96.89 & 97.06 \\
BUCC & ru-en & 97.62 & 97.78 & 95.06 & 96.44 & 45.30 \\
BUCC & zh-en & 97.70 & 99.16 & 95.63 & 97.56 & 97.96 \\
Tatoeba & sqi-eng & 97.22 & 96.76 & 98.17 & 98.57 & 10.38 \\
Tatoeba & fry-eng & 42.07 & 89.31 & 31.13 & 43.54 & 24.62 \\
Tatoeba & kur-eng & 19.09 & 83.59 & 46.94 & 61.44 & 8.26 \\
Tatoeba & tur-eng & 98.03 & 98.00 & 95.08 & 96.17 & 6.15 \\
Tatoeba & deu-eng & 99.07 & 99.20 & 97.02 & 97.73 & 70.10 \\
Tatoeba & nld-eng & 95.35 & 96.07 & 94.58 & 95.50 & 29.74 \\
Tatoeba & ron-eng & 96.52 & 96.92 & 95.30 & 96.43 & 27.23 \\
Tatoeba & ang-eng & 25.22 & 59.28 & 10.24 & 16.72 & 28.76 \\
Tatoeba & ido-eng & 80.86 & 89.42 & 40.25 & 43.91 & 43.91 \\
Tatoeba & jav-eng & 9.95 & 79.77 & 17.04 & 23.39 & 15.02 \\
Tatoeba & isl-eng & 94.32 & 94.75 & 24.07 & 59.25 & 6.29 \\
Tatoeba & slv-eng & 95.40 & 96.03 & 96.92 & 97.08 & 10.14 \\
Tatoeba & cym-eng & 5.85 & 92.00 & 13.25 & 22.31 & 6.97 \\
Tatoeba & kaz-eng & 53.30 & 87.49 & 34.89 & 61.49 & 3.32 \\
Tatoeba & est-eng & 96.43 & 96.55 & 97.33 & 98.40 & 4.76 \\
Tatoeba & heb-eng & 0.00 & 91.53 & 86.88 & 88.26 & 1.69 \\
Tatoeba & gla-eng & 1.52 & 85.66 & 3.61 & 4.72 & 2.09 \\
Tatoeba & mar-eng & 92.93 & 92.65 & 92.38 & 93.83 & 45.53 \\
Tatoeba & lat-eng & 64.81 & 80.07 & 19.47 & 24.25 & 28.76 \\
Tatoeba & bel-eng & 79.54 & 95.00 & 67.73 & 79.94 & 8.03 \\
Tatoeba & pms-eng & 36.23 & 64.57 & 30.70 & 34.19 & 31.94 \\
Tatoeba & gle-eng & 4.20 & 93.80 & 11.62 & 16.85 & 3.26 \\
Tatoeba & pes-eng & 93.13 & 94.70 & 92.59 & 93.47 & 12.13 \\
Tatoeba & nob-eng & 95.77 & 98.40 & 97.73 & 98.53 & 21.07 \\
Tatoeba & bul-eng & 93.57 & 94.58 & 92.65 & 93.52 & 20.09 \\
Tatoeba & cbk-eng & 77.17 & 79.44 & 55.37 & 58.68 & 64.63 \\
Tatoeba & hun-eng & 95.20 & 96.55 & 91.58 & 94.18 & 5.07 \\
Tatoeba & uig-eng & 56.49 & 92.40 & 24.39 & 48.35 & 1.27 \\
Tatoeba & rus-eng & 92.58 & 93.75 & 91.87 & 92.92 & 59.84 \\
Tatoeba & spa-eng & 97.33 & 98.40 & 95.42 & 97.00 & 94.48 \\
Tatoeba & hye-eng & 88.72 & 94.09 & 93.28 & 94.38 & 0.50 \\
Tatoeba & tel-eng & 96.72 & 97.86 & 36.40 & 79.73 & 64.62 \\
Tatoeba & afr-eng & 92.59 & 96.18 & 58.22 & 72.96 & 16.62 \\
Tatoeba & mon-eng & 3.42 & 95.91 & 95.04 & 96.14 & 2.85 \\
Tatoeba & arz-eng & 66.16 & 76.00 & 51.26 & 55.69 & 70.66 \\
Tatoeba & hrv-eng & 96.72 & 96.95 & 95.98 & 97.00 & 12.79 \\
Tatoeba & nov-eng & 60.02 & 74.38 & 47.99 & 50.23 & 52.23 \\
Tatoeba & gsw-eng & 27.52 & 46.50 & 25.74 & 25.12 & 21.03 \\
Tatoeba & nds-eng & 77.13 & 79.42 & 32.16 & 38.88 & 23.92 \\
Tatoeba & ukr-eng & 93.52 & 93.97 & 92.82 & 92.67 & 22.06 \\
Tatoeba & uzb-eng & 23.20 & 84.23 & 17.14 & 23.19 & 4.71 \\
Tatoeba & lit-eng & 96.20 & 96.47 & 93.16 & 95.37 & 4.49 \\
Tatoeba & ina-eng & 93.93 & 95.37 & 79.13 & 84.32 & 73.67 \\
Tatoeba & lfn-eng & 63.39 & 67.54 & 47.02 & 49.56 & 44.85 \\
Tatoeba & zsm-eng & 95.41 & 95.62 & 95.31 & 95.80 & 79.95 \\
Tatoeba & ita-eng & 94.32 & 92.72 & 93.05 & 93.76 & 65.04 \\
Tatoeba & cmn-eng & 85.62 & 95.10 & 94.93 & 95.83 & 91.45 \\
Tatoeba & lvs-eng & 95.33 & 95.88 & 97.87 & 97.53 & 6.55 \\
Tatoeba & glg-eng & 96.14 & 96.82 & 94.00 & 95.32 & 79.86 \\
Tatoeba & ceb-eng & 9.93 & 64.42 & 8.05 & 7.39 & 6.64 \\
Tatoeba & bre-eng & 31.2 & 15.07 & 5.56 & 6.42 & 4.67 \\
Tatoeba & ben-eng & 89.43 & 88.55 & 36.48 & 64.90 & 75.98 \\
Tatoeba & swg-eng & 33.10 & 59.36 & 26.31 & 22.80 & 16.89 \\
Tatoeba & arq-eng & 26.63 & 42.69 & 18.60 & 19.84 & 27.75 \\
Tatoeba & kab-eng & 65.88 & 4.31 & 1.16 & 1.41 & 1.69 \\
Tatoeba & fra-eng & 94.28 & 94.86 & 91.72 & 93.12 & 91.44 \\
Tatoeba & por-eng & 94.54 & 94.14 & 92.13 & 93.02 & 92.62 \\
Tatoeba & tat-eng & 34.74 & 85.92 & 10.25 & 10.89 & 3.59 \\
Tatoeba & oci-eng & 58.13 & 65.81 & 38.57 & 43.49 & 40.17 \\
Tatoeba & pol-eng & 97.32 & 97.22 & 94.28 & 96.95 & 14.09 \\
Tatoeba & war-eng & 8.25 & 60.29 & 7.25 & 7.42 & 10.38 \\
Tatoeba & aze-eng & 82.41 & 94.93 & 62.10 & 76.36 & 6.32 \\
Tatoeba & vie-eng & 96.73 & 97.20 & 95.12 & 97.23 & 94.20 \\
Tatoeba & nno-eng & 72.75 & 94.48 & 76.34 & 81.41 & 16.28 \\
Tatoeba & cha-eng & 14.86 & 31.77 & 15.98 & 12.59 & 23.26 \\
Tatoeba & mhr-eng & 6.86 & 15.74 & 6.89 & 7.57 & 1.56 \\
Tatoeba & dan-eng & 95.22 & 95.71 & 94.80 & 96.17 & 23.52 \\
Tatoeba & ell-eng & 96.20 & 95.35 & 95.43 & 94.93 & 5.34 \\
Tatoeba & amh-eng & 80.82 & 91.47 & 36.21 & 53.49 & 0.03 \\
Tatoeba & pam-eng & 3.24 & 10.73 & 5.41 & 5.39 & 5.85 \\
Tatoeba & hsb-eng & 45.75 & 67.11 & 36.10 & 44.32 & 9.68 \\
Tatoeba & srp-eng & 93.64 & 94.43 & 92.24 & 94.12 & 11.69 \\
Tatoeba & epo-eng & 96.61 & 98.20 & 41.73 & 55.12 & 26.20 \\
Tatoeba & kzj-eng & 4.46 & 11.33 & 6.24 & 5.88 & 5.17 \\
Tatoeba & awa-eng & 33.74 & 71.70 & 33.43 & 42.83 & 35.01 \\
Tatoeba & fao-eng & 57.04 & 87.40 & 27.51 & 38.24 & 12.61 \\
Tatoeba & mal-eng & 98.16 & 98.45 & 32.20 & 88.46 & 83.30 \\
Tatoeba & ile-eng & 87.88 & 85.58 & 57.71 & 60.36 & 59.59 \\
Tatoeba & bos-eng & 95.86 & 94.92 & 93.27 & 94.02 & 13.65 \\
Tatoeba & cor-eng & 4.45 & 10.11 & 3.42 & 3.53 & 2.83 \\
Tatoeba & cat-eng & 95.80 & 95.38 & 94.42 & 96.05 & 88.31 \\
Tatoeba & eus-eng & 93.32 & 95.01 & 23.18 & 31.33 & 53.38 \\
Tatoeba & yue-eng & 87.75 & 89.58 & 71.45 & 77.58 & 77.03 \\
Tatoeba & swe-eng & 95.31 & 95.63 & 94.42 & 95.45 & 19.53 \\
Tatoeba & dtp-eng & 7.39 & 10.85 & 5.69 & 5.03 & 3.41 \\
Tatoeba & kat-eng & 81.16 & 95.02 & 95.44 & 95.46 & 0.42 \\
Tatoeba & jpn-eng & 93.78 & 95.38 & 90.41 & 92.51 & 71.36 \\
Tatoeba & csb-eng & 27.03 & 52.57 & 21.56 & 23.73 & 10.03 \\
Tatoeba & xho-eng & 4.68 & 91.55 & 4.52 & 6.53 & 5.51 \\
Tatoeba & orv-eng & 23.24 & 38.93 & 15.10 & 23.77 & 5.79 \\
Tatoeba & ind-eng & 92.98 & 93.66 & 92.74 & 93.50 & 88.04 \\
Tatoeba & tuk-eng & 16.35 & 75.27 & 15.16 & 14.91 & 5.48 \\
Tatoeba & max-eng & 36.96 & 63.26 & 45.25 & 48.77 & 36.14 \\
Tatoeba & swh-eng & 55.66 & 84.50 & 14.48 & 16.02 & 16.74 \\
Tatoeba & hin-eng & 95.32 & 96.87 & 97.62 & 97.75 & 85.23 \\
Tatoeba & dsb-eng & 42.34 & 64.81 & 33.43 & 36.85 & 8.78 \\
Tatoeba & ber-eng & 77.63 & 8.40 & 4.43 & 4.88 & 4.92 \\
Tatoeba & tam-eng & 87.32 & 89.0 & 24.64 & 73.60 & 72.76 \\
Tatoeba & slk-eng & 95.82 & 96.5 & 95.15 & 96.62 & 9.98 \\
Tatoeba & tgl-eng & 63.19 & 96.02 & 13.09 & 17.67 & 10.70 \\
Tatoeba & ast-eng & 76.35 & 90.68 & 62.17 & 70.08 & 71.13 \\
Tatoeba & mkd-eng & 93.63 & 93.6 & 91.00 & 93.02 & 10.47 \\
Tatoeba & khm-eng & 74.19 & 78.37 & 32.11 & 58.80 & 0.37 \\
Tatoeba & ces-eng & 95.52 & 96.68 & 95.12 & 95.73 & 9.55 \\
Tatoeba & tzl-eng & 36.56 & 58.88 & 25.46 & 34.21 & 27.82 \\
Tatoeba & urd-eng & 84.23 & 93.22 & 94.57 & 95.12 & 70.10 \\
Tatoeba & ara-eng & 90.14 & 88.80 & 87.93 & 90.19 & 85.37 \\
Tatoeba & kor-eng & 87.97 & 90.95 & 92.52 & 93.07 & 22.39 \\
Tatoeba & yid-eng & 2.49 & 88.79 & 14.38 & 30.73 & 0.16 \\
Tatoeba & fin-eng & 96.98 & 96.37 & 93.10 & 95.92 & 3.41 \\
Tatoeba & tha-eng & 96.38 & 96.14 & 96.72 & 95.99 & 2.22 \\
Tatoeba & wuu-eng & 75.09 & 90.18 & 76.00 & 78.25 & 79.58 \\
\midrule
\midrule
Average & mix & 67.42 & 81.75 & 57.98 & 63.38 & 31.08 \\
\hline
    \bottomrule
    \end{tabular}}
    \caption{Multilingual bitext mining results. Scores are f1.}
    \label{tab:addresultsmult}
\end{table*}

\begin{table*}[t!]
    \centering
    \resizebox{0.62\textwidth}{!}{\begin{tabular}{lc|cccccccc}
    \toprule
Dataset & Language & LASER2 & LaBSE & MiniLM-L12-multilingual & MPNet-multilingual & SGPT-BLOOM-7.1B-msmarco \\
\midrule
\midrule
AmazonCounterfactualClassification & de & 67.82 & 73.17 & 68.35 & 69.95 & 61.35 \\
AmazonCounterfactualClassification & ja & 68.76 & 76.42 & 63.45 & 69.79 & 58.23 \\
AmazonReviewsClassification & de & 31.07 & 39.92 & 35.91 & 39.52 & 29.70 \\
AmazonReviewsClassification & es & 32.72 & 39.39 & 37.49 & 39.99 & 35.97 \\
AmazonReviewsClassification & fr & 31.12 & 38.52 & 35.30 & 39.00 & 35.92 \\
AmazonReviewsClassification & ja & 28.94 & 36.44 & 33.24 & 36.64 & 27.64 \\
AmazonReviewsClassification & zh & 30.89 & 36.45 & 35.26 & 37.74 & 32.63 \\
MassiveIntentClassification & af & 38.01 & 56.12 & 45.88 & 52.32 & 47.85 \\
MassiveIntentClassification & am & 12.70 & 55.71 & 36.75 & 41.55 & 33.30 \\
MassiveIntentClassification & ar & 37.16 & 50.86 & 45.14 & 51.43 & 59.25 \\
MassiveIntentClassification & az & 19.98 & 58.97 & 47.42 & 56.98 & 45.24 \\
MassiveIntentClassification & bn & 42.51 & 58.22 & 35.34 & 48.79 & 61.59 \\
MassiveIntentClassification & cy & 17.33 & 50.16 & 26.12 & 27.87 & 44.92 \\
MassiveIntentClassification & da & 45.61 & 58.25 & 57.73 & 62.77 & 51.23 \\
MassiveIntentClassification & de & 44.79 & 56.21 & 50.71 & 59.57 & 56.10 \\
MassiveIntentClassification & el & 46.71 & 57.03 & 58.70 & 62.62 & 46.13 \\
MassiveIntentClassification & es & 45.44 & 58.32 & 59.66 & 64.43 & 66.35 \\
MassiveIntentClassification & fa & 45.01 & 62.33 & 61.02 & 65.34 & 51.20 \\
MassiveIntentClassification & fi & 45.94 & 60.12 & 57.54 & 62.28 & 45.33 \\
MassiveIntentClassification & fr & 46.13 & 60.47 & 60.25 & 64.82 & 66.95 \\
MassiveIntentClassification & he & 42.55 & 56.55 & 52.51 & 58.21 & 43.18 \\
MassiveIntentClassification & hi & 40.20 & 59.40 & 58.37 & 62.77 & 63.54 \\
MassiveIntentClassification & hu & 42.77 & 59.52 & 60.41 & 63.87 & 44.73 \\
MassiveIntentClassification & hy & 28.07 & 56.20 & 51.60 & 57.74 & 38.13 \\
MassiveIntentClassification & id & 45.81 & 61.12 & 59.85 & 65.43 & 64.06 \\
MassiveIntentClassification & is & 39.86 & 54.90 & 30.83 & 37.05 & 44.35 \\
MassiveIntentClassification & it & 48.25 & 59.83 & 59.61 & 64.68 & 60.77 \\
MassiveIntentClassification & ja & 45.30 & 63.11 & 60.89 & 63.74 & 61.22 \\
MassiveIntentClassification & jv & 24.30 & 50.98 & 32.37 & 36.49 & 50.94 \\
MassiveIntentClassification & ka & 22.70 & 48.35 & 43.03 & 49.85 & 33.84 \\
MassiveIntentClassification & km & 22.48 & 48.55 & 40.04 & 45.47 & 37.34 \\
MassiveIntentClassification & kn & 4.32 & 56.24 & 40.98 & 50.63 & 53.54 \\
MassiveIntentClassification & ko & 44.26 & 60.99 & 50.30 & 61.82 & 53.36 \\
MassiveIntentClassification & lv & 39.75 & 57.10 & 54.68 & 61.29 & 46.50 \\
MassiveIntentClassification & ml & 41.33 & 57.91 & 42.41 & 54.34 & 58.27 \\
MassiveIntentClassification & mn & 16.20 & 58.50 & 51.77 & 56.59 & 40.28 \\
MassiveIntentClassification & ms & 43.23 & 58.60 & 54.76 & 60.70 & 59.65 \\
MassiveIntentClassification & my & 25.37 & 57.35 & 52.01 & 57.09 & 37.42 \\
MassiveIntentClassification & nb & 37.74 & 57.91 & 55.50 & 62.60 & 49.41 \\
MassiveIntentClassification & nl & 45.00 & 59.37 & 59.51 & 63.57 & 52.09 \\
MassiveIntentClassification & pl & 44.99 & 59.71 & 59.43 & 64.30 & 50.48 \\
MassiveIntentClassification & pt & 48.55 & 60.16 & 61.27 & 64.89 & 66.69 \\
MassiveIntentClassification & ro & 44.30 & 57.92 & 58.39 & 62.80 & 50.53 \\
MassiveIntentClassification & ru & 44.29 & 60.67 & 59.04 & 63.26 & 58.32 \\
MassiveIntentClassification & sl & 44.72 & 59.37 & 57.36 & 63.51 & 47.74 \\
MassiveIntentClassification & sq & 46.12 & 58.03 & 56.59 & 62.49 & 48.94 \\
MassiveIntentClassification & sv & 45.95 & 59.66 & 59.43 & 64.73 & 50.79 \\
MassiveIntentClassification & sw & 31.89 & 51.62 & 29.57 & 31.95 & 49.81 \\
MassiveIntentClassification & ta & 29.63 & 55.04 & 36.77 & 50.17 & 56.40 \\
MassiveIntentClassification & te & 36.03 & 58.32 & 40.72 & 52.82 & 54.71 \\
MassiveIntentClassification & th & 43.39 & 56.58 & 58.97 & 61.11 & 44.43 \\
MassiveIntentClassification & tl & 29.73 & 55.28 & 33.67 & 38.83 & 50.21 \\
MassiveIntentClassification & tr & 43.93 & 60.91 & 59.90 & 64.54 & 46.56 \\
MassiveIntentClassification & ur & 26.11 & 56.70 & 52.80 & 56.37 & 56.75 \\
MassiveIntentClassification & vi & 44.33 & 56.67 & 56.61 & 59.68 & 64.53 \\
MassiveIntentClassification & zh-CN & 40.62 & 63.86 & 61.99 & 65.33 & 67.07 \\
MassiveIntentClassification & zh-TW & 32.93 & 59.51 & 58.77 & 62.35 & 62.89 \\
MassiveScenarioClassification & af & 47.10 & 63.39 & 53.64 & 59.67 & 51.47 \\
MassiveScenarioClassification & am & 17.70 & 62.02 & 41.89 & 48.97 & 34.87 \\
MassiveScenarioClassification & ar & 45.21 & 57.72 & 51.74 & 57.78 & 65.21 \\
MassiveScenarioClassification & az & 28.21 & 63.48 & 52.06 & 61.53 & 45.58 \\
MassiveScenarioClassification & bn & 50.52 & 61.84 & 41.17 & 54.53 & 67.30 \\
MassiveScenarioClassification & cy & 22.58 & 56.13 & 31.72 & 35.26 & 46.29 \\
MassiveScenarioClassification & da & 54.87 & 65.24 & 66.87 & 71.00 & 53.52 \\
MassiveScenarioClassification & de & 54.34 & 62.39 & 57.40 & 67.34 & 61.74 \\
MassiveScenarioClassification & el & 55.47 & 64.58 & 66.14 & 68.81 & 48.96 \\
MassiveScenarioClassification & es & 52.77 & 63.61 & 65.04 & 70.42 & 73.34 \\
MassiveScenarioClassification & fa & 52.50 & 67.46 & 65.86 & 69.88 & 53.17 \\
MassiveScenarioClassification & fi & 52.63 & 64.58 & 63.75 & 67.60 & 44.69 \\
MassiveScenarioClassification & fr & 54.32 & 65.10 & 66.06 & 70.69 & 72.91 \\
MassiveScenarioClassification & he & 52.41 & 63.53 & 59.20 & 65.16 & 43.10 \\
MassiveScenarioClassification & hi & 47.37 & 64.40 & 65.21 & 67.92 & 69.27 \\
MassiveScenarioClassification & hu & 53.43 & 65.82 & 66.56 & 70.30 & 45.16 \\
MassiveScenarioClassification & hy & 33.57 & 61.25 & 56.11 & 63.02 & 38.73 \\
MassiveScenarioClassification & id & 54.38 & 65.84 & 66.16 & 70.73 & 70.13 \\
MassiveScenarioClassification & is & 49.78 & 61.94 & 37.52 & 44.16 & 44.21 \\
MassiveScenarioClassification & it & 54.84 & 64.09 & 65.00 & 69.73 & 65.57 \\
MassiveScenarioClassification & ja & 54.12 & 67.72 & 66.50 & 69.69 & 65.76 \\
MassiveScenarioClassification & jv & 32.71 & 58.29 & 38.60 & 44.20 & 54.79 \\
MassiveScenarioClassification & ka & 26.92 & 53.38 & 50.66 & 57.30 & 32.99 \\
MassiveScenarioClassification & km & 27.23 & 56.18 & 46.96 & 53.14 & 39.34 \\
MassiveScenarioClassification & kn & 10.06 & 61.74 & 45.73 & 56.08 & 60.50 \\
MassiveScenarioClassification & ko & 52.01 & 67.26 & 55.66 & 68.52 & 55.69 \\
MassiveScenarioClassification & lv & 44.82 & 61.87 & 59.80 & 66.28 & 44.35 \\
MassiveScenarioClassification & ml & 49.10 & 62.26 & 47.69 & 60.13 & 65.53 \\
MassiveScenarioClassification & mn & 21.51 & 62.60 & 57.07 & 60.85 & 38.72 \\
MassiveScenarioClassification & ms & 53.60 & 65.63 & 61.71 & 65.81 & 64.99 \\
MassiveScenarioClassification & my & 29.72 & 62.94 & 59.10 & 63.03 & 36.84 \\
MassiveScenarioClassification & nb & 43.90 & 64.29 & 64.25 & 70.24 & 51.80 \\
MassiveScenarioClassification & nl & 53.33 & 65.16 & 65.52 & 70.37 & 56.32 \\
MassiveScenarioClassification & pl & 52.92 & 64.56 & 65.04 & 68.99 & 49.98 \\
MassiveScenarioClassification & pt & 53.41 & 63.28 & 65.79 & 70.09 & 71.46 \\
MassiveScenarioClassification & ro & 50.48 & 62.41 & 64.17 & 67.95 & 53.69 \\
MassiveScenarioClassification & ru & 51.84 & 65.25 & 65.24 & 69.92 & 61.60 \\
MassiveScenarioClassification & sl & 51.29 & 64.25 & 64.01 & 70.81 & 48.04 \\
MassiveScenarioClassification & sq & 55.65 & 64.54 & 64.31 & 69.63 & 50.06 \\
MassiveScenarioClassification & sv & 54.64 & 66.01 & 67.14 & 71.60 & 51.73 \\
MassiveScenarioClassification & sw & 42.04 & 58.36 & 34.86 & 37.29 & 54.22 \\
MassiveScenarioClassification & ta & 36.72 & 59.08 & 42.62 & 55.96 & 62.77 \\
MassiveScenarioClassification & te & 42.08 & 64.13 & 46.46 & 58.81 & 62.59 \\
MassiveScenarioClassification & th & 52.15 & 64.34 & 67.01 & 69.44 & 45.18 \\
MassiveScenarioClassification & tl & 37.34 & 60.23 & 37.37 & 43.99 & 52.06 \\
MassiveScenarioClassification & tr & 52.56 & 65.43 & 66.55 & 70.4 & 47.21 \\
MassiveScenarioClassification & ur & 32.60 & 61.52 & 60.43 & 62.9 & 64.26 \\
MassiveScenarioClassification & vi & 50.97 & 61.05 & 60.72 & 65.71 & 70.61 \\
MassiveScenarioClassification & zh-CN & 50.22 & 70.85 & 67.44 & 71.23 & 73.95 \\
MassiveScenarioClassification & zh-TW & 42.32 & 67.08 & 65.70 & 68.73 & 70.30 \\
MTOPDomainClassification & de & 74.08 & 86.95 & 79.20 & 85.73 & 82.05 \\
MTOPDomainClassification & es & 73.47 & 84.07 & 83.04 & 86.96 & 93.55 \\
MTOPDomainClassification & fr & 72.26 & 84.14 & 78.63 & 81.21 & 90.98 \\
MTOPDomainClassification & hi & 72.95 & 85.11 & 81.36 & 84.76 & 89.33 \\
MTOPDomainClassification & th & 72.68 & 81.24 & 79.99 & 82.51 & 60.49 \\
MTOPIntentClassification & de & 51.62 & 63.42 & 54.23 & 61.27 & 61.92 \\
MTOPIntentClassification & es & 52.75 & 64.44 & 60.28 & 66.59 & 74.49 \\
MTOPIntentClassification & fr & 50.12 & 62.01 & 54.05 & 59.76 & 69.12 \\
MTOPIntentClassification & hi & 45.55 & 62.58 & 59.90 & 62.37 & 64.85 \\
MTOPIntentClassification & th & 50.07 & 64.61 & 61.96 & 64.80 & 49.36 \\
\midrule
\midrule
Average & mix & 42.85 & 60.77 & 54.87 & 60.39 & 54.4 \\
   \bottomrule
    \end{tabular}}
    \caption{Multilingual classification results. Scores are accuracy.}
    \label{tab:addresultsmultclf}
\end{table*}

\begin{table*}[t!]
    \centering
    \resizebox{\textwidth}{!}{\begin{tabular}{ll|ccccccc}
    \toprule
Dataset & Language & Komninos & LASER2 & LaBSE & MiniLM-L12-multilingual & MPNet-multilingual & SGPT-BLOOM-7.1B-msmarco \\
\midrule
\midrule
STS17 & ko-ko & 2.54 & 70.52 & 71.32 & 77.03 & 83.41 & 66.89 \\
STS17 & ar-ar & 13.78 & 67.47 & 69.07 & 79.16 & 79.10 & 76.42 \\
STS17 & en-ar & 9.08 & 65.05 & 74.51 & 81.22 & 80.85 & 78.07 \\
STS17 & en-de & -3.11 & 66.66 & 73.85 & 84.22 & 83.28 & 59.10 \\
STS17 & en-tr & -0.45 & 70.05 & 72.07 & 76.74 & 74.90 & 11.80 \\
STS17 & es-en & -8.18 & 55.30 & 65.71 & 84.44 & 86.11 & 78.22 \\
STS17 & es-es & 48.23 & 79.67 & 80.83 & 85.56 & 85.14 & 86.00 \\
STS17 & fr-en & 5.81 & 70.82 & 76.98 & 76.59 & 81.17 & 80.46 \\
STS17 & it-en & 3.64 & 70.98 & 76.99 & 82.35 & 84.24 & 51.58 \\
STS17 & nl-en & -0.44 & 68.12 & 75.22 & 81.71 & 82.51 & 45.85 \\
STS22 & de & 33.04 & 25.69 & 48.58 & 44.64 & 46.70 & 30.05 \\
STS22 & es & 48.53 & 54.92 & 63.18 & 56.56 & 59.91 & 65.41 \\
STS22 & pl & 12.47 & 18.34 & 39.30 & 33.74 & 33.65 & 31.13 \\
STS22 & tr & 47.38 & 36.97 & 58.15 & 53.39 & 56.30 & 47.14 \\
STS22 & ar & 32.42 & 42.57 & 57.67 & 46.2 & 52.19 & 58.67 \\
STS22 & ru & 19.44 & 39.24 & 57.49 & 57.08 & 58.74 & 43.36 \\
STS22 & zh & 4.78 & 49.41 & 63.02 & 58.75 & 61.75 & 66.78 \\
STS22 & fr & 49.43 & 58.61 & 77.95 & 70.55 & 74.30 & 80.38 \\
STS22 & de-en & 28.65 & 32.35 & 50.14 & 52.65 & 50.81 & 51.16 \\
STS22 & es-en & 26.97 & 54.34 & 71.86 & 67.33 & 70.26 & 75.06 \\
STS22 & it & 57.77 & 60.31 & 72.22 & 55.22 & 60.65 & 65.65 \\
STS22 & pl-en & 45.55 & 53.63 & 69.41 & 69.02 & 73.07 & 53.31 \\
STS22 & zh-en & 14.05 & 46.19 & 64.02 & 65.71 & 67.96 & 68.45 \\
STS22 & es-it & 41.10 & 42.21 & 69.69 & 47.67 & 53.70 & 65.50 \\
STS22 & de-fr & 14.77 & 37.41 & 53.28 & 51.73 & 62.34 & 53.28 \\
STS22 & de-pl & 11.21 & 15.67 & 58.69 & 44.22 & 40.53 & 43.05 \\
STS22 & fr-pl & 39.44 & 39.44 & 61.98 & 50.71 & 84.52 & 28.17 \\
\midrule
\midrule
Average & mix & 22.14 & 51.55 & 65.67 & 64.23 & 67.71 & 57.81 \\
    \bottomrule
    \end{tabular}}
    \caption{Multilingual STS Results. Scores are Spearman correlations of cosine similarities.}
    \label{tab:addresultsmultsts}
\end{table*}

\end{document}